\documentclass[sigconf,nonacm]{aamas} 
\usepackage[utf8]{inputenc}

\usepackage{balance} % for balancing columns on the final page

\usepackage{algorithm}
\usepackage{algorithmic}
\usepackage{graphicx}
\usepackage{textcomp}
\usepackage{siunitx}
\usepackage{xcolor}
\usepackage{subcaption}
\usepackage{wrapfig}

\usepackage[
raggedright
]{sidecap}   
\sidecaptionvpos{figure}{t} 
\makeatletter
\gdef\@copyrightpermission{
  \begin{minipage}{0.2\columnwidth}
   \href{https://creativecommons.org/licenses/by/4.0/}{\includegraphics[width=0.90\textwidth]{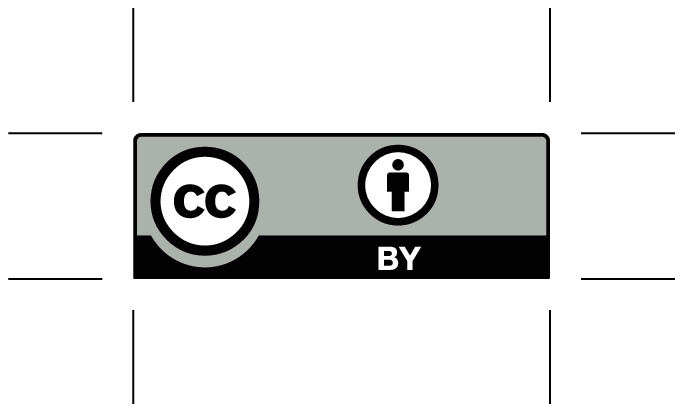}}
  \end{minipage}\hfill
  \begin{minipage}{0.8\columnwidth}
   \href{https://creativecommons.org/licenses/by/4.0/}{This work is licensed under a Creative Commons Attribution International 4.0 License.}
  \end{minipage}
  \vspace{5pt}
}
\acmConference[]{}{}{}{}
\copyrightyear{}
\acmYear{}
\acmDOI{}
\acmPrice{}
\acmISBN{}

\definecolor{valid}{rgb}{0., 0.8, 0.}

\newcommand{\R}{\mathbb{R}}
\newcommand{\Distrib}{\mathcal{D}}
\newcommand{\proba}{\mathbb{P}}
\newcommand{\buffer}{\mathcal{B}}
\newcommand{\probSpace}{\mathcal{P}}
\newcommand{\actionSpace}{\mathcal{A}}
\newcommand{\stateSpace}{\mathcal{S}}
\newcommand{\goalSpace}{\mathcal{G}}
\newcommand{\rGC}{r_{\mathcal{G}}}
\newcommand{\pG}{p_{\mathcal{G}}}
\newcommand{\rS}{r_S}
\newcommand{\expect}{\mathbb{E}}
\newcommand{\aS}{a_S}
\newcommand{\aGC}{a_{GC}}
\newcommand{\piS}{\pi_{\phi_S}}
\newcommand{\piGC}{\pi_{\phi_{GC}}}
\newcommand{\thGCtoS}{\text{th}_{GC \rightarrow S}}
\newcommand{\thStoGC}{\text{th}_{S \rightarrow GC}}
\newcommand{\SafetyActivated}{\text{safety}\_\text{flag}}
\newcommand{\neighborhood}{N_0}

\newcommand{\refAlg}[1]{Algorithm \ref{#1}}
\newcommand{\refFig}[1]{Figure \ref{#1}}
\newcommand{\refEq}[1]{(\ref{#1})}
\newcommand{\refSection}[1]{section \ref{#1}}
\acmSubmissionID{1042}

\title[AAMAS-2025 Formatting Instructions]{Learning to explore when mistakes are not allowed}

\author{Charly Pecqueux-Guézénec}
\affiliation{
  \institution{Sorbonne Université, CNRS, ISIR}%\institutenfrancais
  \city{F-75005 Paris}
  \country{France}}
\email{pecqueuxguezenec@isir.upmc.fr} % TODO: google or lab ? 
\author{Stéphane Doncieux}
\affiliation{
  \institution{Sorbonne Université, CNRS, ISIR}%\institutenfrancais
  \city{F-75005 Paris}
  \country{France}}
\email{doncieux@isir.upmc.fr}
\author{Nicolas Perrin-Gilbert}
\affiliation{
  \institution{Sorbonne Université, CNRS, ISIR}%\institutenfrancais
  \city{F-75005 Paris}
  \country{France}}
\email{perrin@isir.upmc.fr}
\begin{abstract}
Goal-Conditioned Reinforcement Learning (GCRL) provides a versatile framework for developing 
unified controllers capable of handling wide ranges of tasks, exploring environments, 
and adapting behaviors. However, its reliance on trial-and-error poses challenges for real-world 
applications, as errors can result in costly and potentially damaging consequences. To address the 
need for safer learning, we propose a method that enables agents to learn goal-conditioned behaviors 
that explore without the risk of making harmful mistakes. Exploration without risks can seem paradoxical, 
but environment dynamics are often uniform in space, therefore a policy trained for safety without 
exploration purposes can still be exploited globally. Our proposed approach involves two distinct 
phases. First, during a pretraining phase, we employ safe reinforcement learning and distributional
techniques to train a safety policy that actively tries to avoid failures in various situations. 
In the subsequent safe exploration phase, a goal-conditioned (GC) policy is learned while ensuring safety. 
To achieve this, we implement an action-selection mechanism leveraging the previously learned 
distributional safety critics to arbitrate between the safety policy and the GC policy, ensuring 
safe exploration by switching to the safety policy when needed.
We evaluate our method in simulated environments and demonstrate that it not only provides 
substantial coverage of the goal space but also reduces the occurrence of mistakes to a minimum, 
in stark contrast to traditional GCRL approaches. Additionally, we conduct an ablation study and 
analyze failure modes, offering insights for future research directions.

\end{abstract}

\keywords{Safe Exploration, Safe Reinforcement Learning, Goal-Conditioned Reinforcement Learning}
\newcommand{\BibTeX}{\rm B\kern-.05em{\sc i\kern-.025em b}\kern-.08em\TeX}

\begin{document}

%%% The following commands remove the headers in your paper. For final 
%%% papers, these will be inserted during the pagination process.

\pagestyle{fancy}
\fancyhead{}

%%% The next command prints the information defined in the preamble.

\maketitle 

%%%%%%%%%%%%%%%%%%%%%%%%%%%%%%%%%%%%%%%%%%%%%%%%%%%%%%%%%%%%%%%%%%%%%%%%

\section{Introduction}

Goal-conditioned reinforcement learning (GCRL) provides 
bases to build a single robotic controller that can achieve a wide variety of tasks and
adapt its behavior to its environment \cite{colas2022autotelic, VGCRL, UVFA, HER}.
GCRL methods have shown impressive performance on a wide variety of tasks, in simulation but also in the real world, 
essentially with manipulator robots \cite{RIS,skewfit,PlanningWithGCpolicies}.
In the case of manipulator robots, the action is often the desired position of the end effector. Then the agent strongly depends on primitives, 
facilitating exploration and avoiding safety problems.  

However, moving towards fully autonomous agents, we would like to build robots that can explore their environment 
and discover new skills while staying out of danger. An example of such an ideal case would be 
a humanoid robot that does not fall while learning to move around in a new environment. Though it may seem contradictory at first glance, 
humans constantly explore new environments with the confidence that their balance and walking skills will enable them to do it safely.
Hikers, for instance, can navigate in unknown terrains with minor risks as 
their knowledge on how to avoid falling is general and applicable to diverse environments. 
Transposing these intuitions into formal vocabulary from viability theory \cite{viability_wieber}, hikers know how to remain in viable states, even when they explore.

\begin{figure}[ht!]
    \centering
    \includegraphics[width=0.45\textwidth]{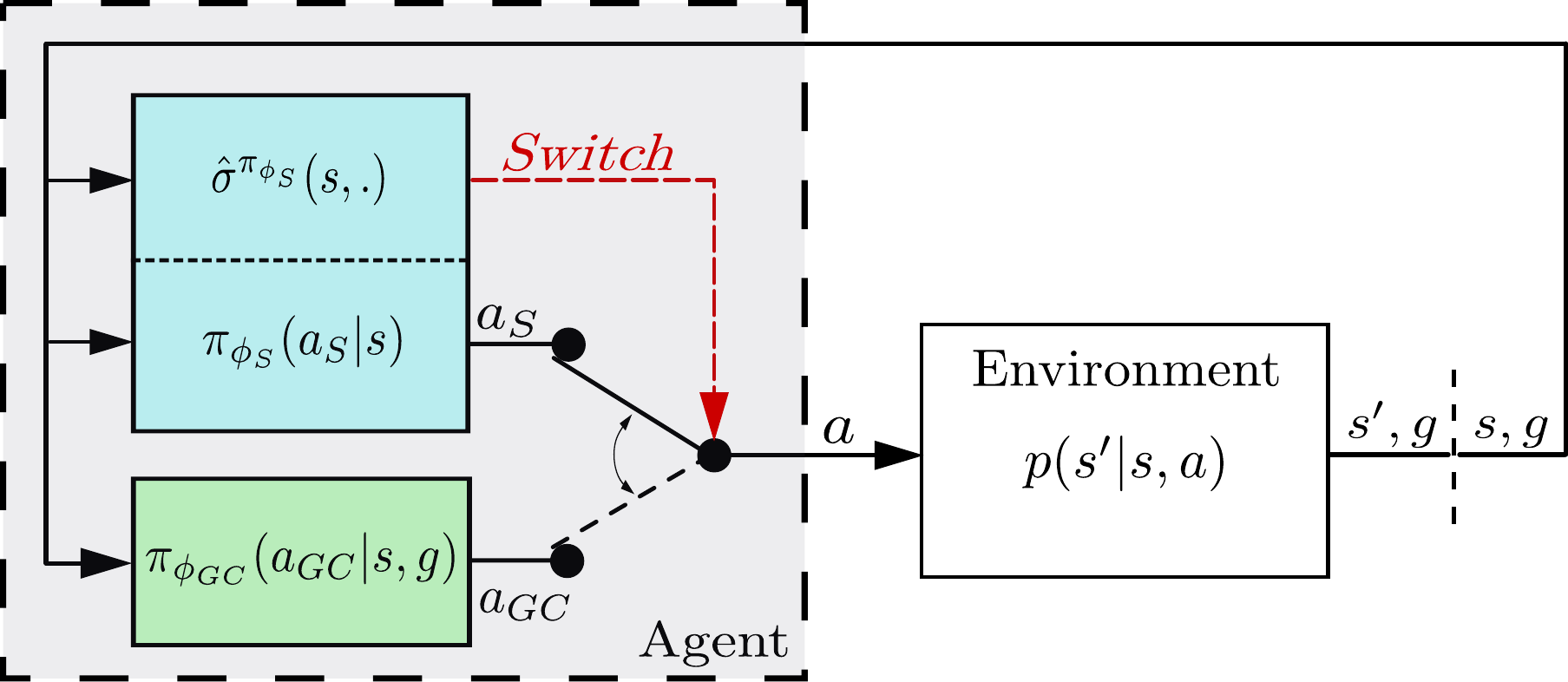}
    \caption{Action selection mechanism to guarantee safe exploration. The agent 
    observes the current state $s$ of the environment and the current goal $g$.
    The safety policy samples an action $a_S \sim \piS(.|s)$ that must prevent future mistakes, while the 
    goal-conditioned policy samples an action $a_{GC} \sim \piGC(.|s,g)$ to go towards $g$. 
    For each possibility, the function $\sigma^{\piS}$ estimates the level of confidence in the safety policy's ability to avoid potential future errors. If it is too low, the safe action $a_S$ is executed to keep the system safe. Otherwise, $a_{GC}$ is executed, allowing the agent to explore. 
    }
    
    \Description{Action selection mechanism to guarantee safe exploration. The agent 
    observes the current state $s$ of the environment and the current goal $g$.
    The safety policy samples an action $a_S \sim \piS(.|s)$ that must prevent future mistakes, while the 
    goal-conditioned policy samples an action $a_{GC} \sim \piGC(.|s,g)$ to go towards $g$. 
    For each possibility, the function $\sigma^{\piS}$ estimates the level of confidence in the safety policy's ability to avoid potential future errors. If it is too low, the safe action $a_S$ is executed to keep the system safe. Otherwise, $a_{GC}$ is executed, allowing the agent to explore. }
    \label{fig:schema_switch}
\end{figure}

We propose a safe exploration method that combines a goal-conditioned (GC) policy aiming to explore, discover and 
learn to cover a goal space of positions, and a safety policy aiming at maintaining the system in viable states. As the system dynamics are usually invariant in space and independent from the agent's goals, the safety policy
is defined independently from goals with the hope that it will be globally reliable. The method involves two distinct phases: a first phase of \textit{pretraining}
and a second phase of \textit{safe exploration}. In the first phase, the safety policy is trained in simulation 
using safe reinforcement learning (safe RL) and distributional critics. In the second phase, the GC policy is
trained. To avoid mistakes during exploration, an action-selection mechanism estimates the risk for future states to be 
unviable thanks to the distributional critics of the safety policy, then chooses which policy to execute at the current step. 
Obtaining theoretical results would depend on strong assumptions about the environment dynamics and near-perfect knowledge from the safety policy. Instead, our work focuses on empirically demonstrating that, in practice, an agent can learn to explore safely without making mistakes.
Contrary to other approaches assuming
to have access to instantaneous emergency actions \cite{MASESafeExp,dalal2018safetyLayer}, 
we test our approach on two environments for which there exist states that 
irreversibly lead to mistakes or failures: a goal-based version of CartPole and an environment based on the Skydio X2 drone from Mujoco Menagerie
\cite{towers2024gymnasium,menagerie2022github}. In these environments, random actions or even zero actions lead to mistakes, which makes the conservation of safety quite challenging. To our knowledge, our approach is the only one that explicitly gathers safe RL and GCRL to perform safe exploration.

Our contributions are threefold:
\begin{itemize}
    \item The design of a distributional safe RL framework to pretrain a safety policy that will prevent mistakes during the exploration phase.
    \item The design of an action selection mechanism to ensure safe exploration.
    \item The study of the key components of the method and failure modes to orient future research.
\end{itemize}

\section{Related Work}
\label{section:related_work}

\paragraph{\textbf{Safe Reinforcement Learning}}
Constrained optimal control methods allow to synthesize controllers with safety guarantees but they assume a perfect or at least partial knowledge of the system dynamics \cite{CBF_th_2019,learningbasedMPCkoller2019,safe_control_certificates_dawson2022safe}. 
On the contrary, model-free safe reinforcement learning (safe RL) only assumes
the system dynamics to be stochastic and markovian. 
Most safe RL methods aim at solving a Constrained Markov Decision Process (CMDP),
which is built upon an analogy between rewards and constraints
\cite{AltmanCMDP,ha2020SACLagLevine,2015safeRLComprehensive,achiam2017CPO,lyapunovbased_policy_gradient_chow2019},
as they formulate the constraint as a discounted cumulative objective, which is convenient 
from the RL point of view but not sufficient to satisfy persistent safety constraints along trajectories.
To overcome this issue, some methods reformulate the critic target. For example, in \cite{2020safetyCritic}, a safety critic allows to formulate the constraint in terms of failure probability \cite{2020safetyCritic}. 
Reachability-constrained RL (RCRL) \cite{RCRL2022} proposes to use a $\max$ operator in the safety critic target to discover 
the largest set of reachable states by the policy.
%Reachability-constrained RL (RCRL) uses the $\max$ operator to obtain the largest feasible set \cite{RCRL2022} {\color{blue} CLARIFIER}. 
In our work, we pre-train a safety policy using ingredients from RCRL and then use the critic in the
safe exploration phase to decide when to switch from one policy to another.
Also, since our goal is to prevent the worst-case scenario,
we use quantile-based distributional reinforcement learning to train our safety policy 
\cite{QR-DQN,WorstCasePG2019,WCSAC2021}.

\paragraph{\textbf{Goal-conditioned RL}}
The standard RL framework addresses a single task, specified by the reward function.
Goal-conditioned reinforcement learning (GCRL) provide 
bases to build a single robotic controller that is able to achieve a wide variety of tasks and
adapt its behavior to its environment \cite{Kaelbling1993LearningTA,UVFA,colas2022autotelic}.
A goal can be a desired position for a given rigid body in the environment. In addition to being convenient 
for a robot user, it allows to learn from failed attempts. Indeed, an agent that has failed to reach
a given desired goal has actually reached another accidentally.
This feature can be exploited by relabelling methods, like Hindsight Experience Replay
(HER), to learn a goal-conditioned policy from sparse rewards. 
In our experiments, we consider a sparse reward setting and combine  
Soft Actor-Critic (SAC) and HER to learn goal-conditioned policies \cite{SAC, HER}. 

\paragraph{\textbf{Learning diverse safe skills}}
\citeauthor{ha2020SACLagLevine} have developed a method to learn safe locomotion in three directions 
for the Minitaur robot. Although policy improvement over time leads to constraint satisfaction, 
reducing the need for humain intervention for resets, the formalism does not explicitly enforce constraint 
satisfaction during exploration \cite{ha2020SACLagLevine}. SASD on the other hand combines 
unsupervised skill discovery and safe RL to learn a richer set of skills, but still does not enforce safety
during the exploration phase \cite{SASD}.

\paragraph{\textbf{Safe exploration}}
\citeauthor{exploration_survey} identify two categories of safe exploration methods: those based on
auxiliary reward and those based on human designer knowledge \cite{exploration_survey}. 
Auxiliary reward methods penalize the agent when it puts itself in danger. 
By definition, these methods incentivize the agent to avoid repeated mistakes and catastrophic behaviors, 
but do not constrain the exploration policy behavior directly. As a result, they reduce the occurrence of 
mistakes compared to baselines \cite{karimpanal2020learning} or even improve learning 
\cite{lipton2018sisyphean, fatemiDeadEnds} but do not prevent them.
In methods based on human designer knowledge, boundaries safe and unsafe states or behaviors 
are defined by the human designer. Most of these methods make strong assumptions about the environment
or the agent behavior, which allows the use of baseline behaviors, 
human intervention, hand-crafted model or heuristics
\cite{GarciaSafeExp2012,VerifSafeExp,TrialWithoutErrorSafeExp,SafeExpGPMDP}. 
A recent approach proposes a Meta-Algorithm for Safe Exploration (MASE) and provides theoretical guarantees on 
optimality and safety during exploration \cite{MASESafeExp}. However, this approach assumes the agent has access to an emergency stop action that can reset the environment when no viable actions are available, an assumption that does not hold in general, and particularly not in our environments. In contrast, our focus is on environments where unviable states exist, and the agent must navigate within these constraints.
Closer to us, Dalal et al. \cite{dalal2018safetyLayer} use a pre-trained linearized constraint model and solve a quadratic program to project actions onto a feasible set during exploration. Their method, however, assumes access to a linearized model and that constraint violations can be avoided in a single step, which is not our case.
Like us, Srinivasan et al. \cite{2020safetyCritic} propose two training phases. But they pre-train and then fine-tune a single policy, while we consider two different policies that are trained separately and, above all, do not share the same input. 
In our approach, the constraint function is used as an auxiliary reward signal for the safety policy, unlike other auxiliary reward methods where it influences the task-solving policy. Additionally, by making general assumptions about the environment's structure, such as the uniformity in space of the dynamics, our method straddles both categories.
Also, contrary to previously cited approaches, we explicitly merge GCRL with safe RL tools to build our safe exploration framework.

%\textcolor{blue}{SAC-N\cite{sac_n_edac} (ATTENTION Finn utilise autre chose)}

%\textcolor{blue}{Safe exploration in parameters space (GP and QD): \cite{SafeOpt,SafeExpPeters,SafeExpActiveLearningGP,ResetFreeQD}}

\section{Background}

In this section, we introduce the notations and building blocks needed to develop our safe exploration method: 
RL and safe RL for learning the safety policy, goal-conditioned RL to achieve goals and distributional critics 
for risk-aware action selection during the safe exploration phase.

\subsection{Reinforcement Learning formalism}
\label{subsection:RL}

An RL problem is described by a tuple $(\stateSpace, \actionSpace, p, p_0, r)$, called a Markov decision process (MDP),
where $\stateSpace$ is the state space, $\actionSpace$ the action space,
$p : \stateSpace \times \actionSpace \to \probSpace(\stateSpace)$ the transition function, 
$p_0 \in \probSpace(\stateSpace)$ the initial state probability distribution, 
$r : \stateSpace \times \actionSpace \to \R$ the reward function. 
The objective in RL is to obtain a policy $\pi : \stateSpace \to \probSpace(\actionSpace)$ 
that maximizes the expected sum of discounted rewards $\expect_{\pi} \left[ \sum_{t=0}^{+\infty} \gamma^t r(s_t, a_t) \right]$, where $\gamma \in [0,1]$ is the discount factor \cite{SuttonBartho}.

\subsection{Goal-conditioned RL formalism}
\label{subsection:GCRL}

Goal-conditioned RL (GCRL) extends the RL framework to a multi-goal setting 
$(\stateSpace \times \goalSpace, \actionSpace, p, p_0, \pG, \rGC)$, where $\goalSpace$ is the goal space, 
\textit{eg} positions to reach, $\rGC : \stateSpace \times \actionSpace \times \goalSpace \to \R$ the 
new reward function, and $\pG \in \probSpace(\goalSpace)$ the distribution of goals sampled at the beginning of each episode.
In this setting, we consider a goal-conditioned (GC) policy $\pi : \stateSpace \times \goalSpace \to \probSpace(\actionSpace)$ 
whose objective is to maximize $\expect_{g\sim\pG, \pi(.|.,g)} \left[ \sum_{t=0}^{+\infty} \gamma^t \rGC(s_t, a_t, g) \right]$, 
so that the resulting policy maximizes goal space coverage \cite{UVFA}.
In our framework, we consider a sparse reward setting, where the reward is equal to $1$ when the goal $g$ is reached 
and $0$ otherwise. It avoids the need for complicated reward engineering but requires relabelling to learn from failed attempts.
We use Hindsight Experience Replay (HER) with "future" strategy combined with SAC for
policy learning \cite{HER, SAC}. 

\subsection{Safe Reinforcement Learning}
\label{subsection:SafeRL}

Safe RL aims at solving an RL problem under constraints, summarized in a Constrained MDP (CMDP) 
$(\stateSpace, \actionSpace, p, p_0, \rS, h)$, where $h: \stateSpace \to \R$ is 
the constraint function \cite{AltmanCMDP}. In many safe RL algorithms $h$ plays the 
same role as a reward function and the constraint is based on the expected sum of discounted costs (Cf \refSection{section:related_work}). 
As the sum form allows compensations between terms, it does not guarantee that every state $s$ in an episode verifies $h(s) \le 0$.
Therefore, we rather opted for the RCRL framework and its reachability critic that offer such a guarantee \cite{RCRL2022}. Also, it relates the reachability value to a notion of distance between the current state and the set of unsafe states, which is a crucial aspect of our action selection mechanism. 

\subsection{Distributional Reinforcement Learning}

Safe exploration is inherently related to a notion of risk. To explore the agent has to go through previously unvisited states.
But to do so it must already have some \textit{a priori} knowledge about the safety of the states it is about to visit.
A balancing robot for example may have to evaluate the probability to fall given its current state $s$ and the action $a$ it is
about to execute. 
Distributional RL methods based on quantile regression, like TQC, allow to compute such a probability \cite{QR-DQN,TQC}. Rather than computing the mean of the return distribution, their critic function approximates the entire distribution using a sum of Dirac delta functions $Z_{\psi}(s, a) := \frac{1}{N}\sum_{i=1}^{N} \delta_{\theta_{\psi}^{(i)}(s, a)}$. Each Dirac position $\theta_{\psi}^{(i)}(s, a)$ corresponds to a quantile of the return distribution, and $N$ is the number of quantiles. Parameter $\psi$ is optimized via quantile regression Q-learning \cite{QR-DQN} updates on batches of transitions uniformly sampled from a replay buffer $\buffer$. The mean of all quantiles is the expected return $\mathbb{E} \left[ \left. \sum_{t=0}^{+\infty} \gamma^t \rS(s_t, a_t) \right| s, a \right]$ and we use it for policy learning. 

By definition, each quantile $\theta_{\psi}^{(i)}(s, a)$ is associated to a cumulative probability $\hat{\tau_i} \triangleq \proba\left(Z^{\pi}(s, a) \le \theta_{\psi}^{(i)}(s, a)\right) = \frac{2.i - 1}{2.N}$, with $i\in\{1,...,N\}$ \cite{QR-DQN}. We assume that the sum of rewards must not fall below a given threshold $v$, which is associated to a potential mistake. As the cumulative distribution function is non-decreasing, if $\theta_{\psi}^{(i)}(s, a) \le v$, then $\hat{\tau_i} = \frac{2.i - 1}{2.N} \le \proba\left(Z^{\pi}(s, a) \le v\right)$. As a result, the relative positions of the quantiles with respect to a given threshold allow us to master the risk level during the safe exploration phase of our method. In our implementation, for a given hyperparameter $\tau \in [0, 1]$, we compute the mean of quantiles $\theta_{\psi}^{(i)}(s, a)$ that verify $\hat{\tau_i} \le \tau$, then we compare it to the threshold $v$ to decide whether switching to the safety policy is necessary. For example, $\tau = 0.1$ corresponds to the worst $10 \%$ of possible value outcomes.

\section{Method}

In this section, we formalize the safe exploration problem and detail the method we propose to solve it.

\subsection{Defining the safe exploration problem}
\label{subsection:safe_exp_problem}

We aim for our agent to learn how to maximize coverage of a goal set, which corresponds to solving a multi-goal MDP (\refSection{subsection:GCRL}). However, the agent must also explore its environment safely, avoiding mistakes. In our framework, a mistake occurs when a terminal state is reached. Therefore, the agent must avoid terminal states during exploration. To do so, we provide our agent with a safety policy that we can activate when the GC policy is about to make a mistake and lead the agent out of danger. 

A way to build this safety policy is under the angle of classic RL. We consider a set $\neighborhood \subset \stateSpace$ of states that are desirable in terms of safety, for instance near some equilibrium point. The safety reward $\rS$ equals $1$ if $s \in \neighborhood$ and 0 otherwise. This reward setting is convenient as it relates the sum of rewards to the number of steps $T^{\piS}(s, a)$ necessary to reach $\neighborhood$ from a given state-action couple $(s, a)$. Then, to decide when to switch from one policy to the other we can compare the estimation given by the safety critic of the number of steps necessary to reach $ \neighborhood$ to a threshold.

However, critics are neural networks, which are continuous functions, while the number of steps is finite for states from which we can reach $\neighborhood$ and infinite for terminal states regarding an optimal safety policy. Thus, it pushes the critic to generalize safety to unsafe states and leads the agent to believe that it can still use the GC policy even though objectively, it is about to make a mistake. Therefore, in addition to the temporal distance, we added a notion of distance in the state space between the current state and the set of terminal states. To do so, we assume the agent receives a cost value $h(s)$ at each environment step, where $s$ is the current state and $h: \stateSpace \to \R$ a continuous constraint function that verifies $h(s) > 0$ if and only if $s$ is a terminal state. In our framework, we identify terminal states as mistakes the agent absolutely has to avoid. Like in RCRL, the function $h$ is related to a kind of distance between the current state $s$ and the set of terminal states \cite{RCRL2022}. Assuming to have access to such a function is not very restrictive as robots have sensors and often state estimation modules. The continuity of function $h$ is crucial as it allows for generalization from known states to unseen states. Indeed, an unvisited state near another visited and safe state which is far enough from terminal states is likely to be safe too. On the contrary, states near unsafe states are likely to be unsafe. As a result, we switch from the GC policy to the safety policy if the number of steps to reach $\neighborhood$ is too high or if $h(s)$ is close to $0$. 

All these considerations lead us to define the safe exploration problem as the combination of a CMDP $(\stateSpace, \actionSpace, p, p_0, \rS, h)$
and a multi-goal MDP $(\stateSpace \times \goalSpace, \actionSpace, p, p_0, \pG, \rGC)$:

\begin{equation}
    \mathcal{T} = (\stateSpace \times \goalSpace, \actionSpace, p, p_0, \pG, \rS, \rGC, h)
    \label{eq:SafeExpProblem}
\end{equation}

In the pretraining phase, we train a parametrized stochastic safety policy $\piS$ that solves the CMDP. 
The CMDP is independent of the goal space as the notion of safety is independent of the goals pursued by the agent.
We assume that we have access to simulation to perform the safety pretraining, allowing us to perform reset anywhere 
and then to train the policy on a wide variety of situations. The critics that have been trained are then used in the 
action selection mechanism.

During the safe exploration phase, the safety policy and its critics are fixed and we train a GC policy. 
The multi-goal MDP part follows the framework developed in \refSection{subsection:GCRL}. 
Transitions are collected and stored in an episodic replay buffer $\buffer$, regardless of the policy that generates them.
Thus transitions generated by both policies can be found in the same episode. The main idea behind is that both policies 
can learn from each other as their respective objectives may be complementary in some situations.

\subsection{Safety policy learning}
\label{subsection:safety_policy_learning}

To train the safety policy, we take inspiration from the distributional algorithm TQC, as it uses an ensemble critics for robustness and truncation in the critic update to prevent overestimation \cite{TQC}.
On the one hand, we train $M$ approximations $Z_{\psi_1}, ... Z_{\psi_M}$ of the safety policy return distribution $Z^{\piS}(s, a) = \Distrib_{\piS} \left[ \left. \sum_{t=0}^{+\infty} \gamma^t \rS(s_t, a_t) \right| s, a \right]$ using the TQC critic loss. The targets $Z_{\overline{\psi_1}}, ... Z_{\overline{\psi_M}}$ are initialized in the same way and follow $Z_{\psi_1}, ... Z_{\psi_M}$ via exponential moving average update. 
Similarly, we train an ensemble of $M$ reachability critics $R_{\xi_1}, ... R_{\xi_M}$ but we replace the usual target based on a sum with the RCRL \cite{RCRL2022} target based on a $\max$ operator, leading to equation \refEq{eq:RCRL_target}. 
Likewise, targets $R_{\overline{\xi_1}}, ... R_{\overline{\xi_M}}$ follow $R_{\xi_1}, ... R_{\xi_M}$ via exponential moving average. Unlike TQC, the reachability critics are updated separately using the quantile regression loss \cite{QR-DQN}.
Critic ensemble are trained on the same batched transitions but initialized with different seeds.

\begin{equation}
    \mathcal{T}\theta^{(j)} = (1-\gamma) h(s') + \gamma \max\left( h(s'), \theta_{\overline{\xi}}^{(j)}(s', a') \right)
    \label{eq:RCRL_target}
\end{equation}

Policy parameters $\phi_S$ are optimized to minimize the following loss, also inspired by TQC \cite{TQC}:

\begin{equation}
    \mathcal{L}_{\pi_S}(\phi_S) = \expect_{\substack{s, a \sim \mathcal{B}}}
    \left[ \alpha \log \piS(a|s) - \overline{Q}(s, a) + \lambda \overline{R}(s, a) \right]
    \label{eq:safety_actor_loss}
\end{equation}
where $\overline{Q}(s, a) = \frac{1}{M N} \sum_{i = 1}^{M} \sum_{j = 1}^{N} \theta_{\psi_i}^{(j)}(s, a)$

$\overline{R}(s, a) = \frac{1}{M N} \sum_{i = 1}^{M} \sum_{j = 1}^{N} \theta_{\xi_i}^{(j)}(s, a)$ and $\lambda \in \R_+$ is
a positive multiplier that we keep fixed. In our experiments, we performed an ablation study to identify the effect of 
reachability critics on safety during exploration. So we tested $\lambda = 0$ and $\lambda = 100$. The value of $100$
has been chosen so that $\overline{Q}(s, a)$ and $\overline{R}(s, a)$ have the same scale. We also tested a varying $\lambda$, 
like \citeauthor{ha2020SACLagLevine}, but it led to too much instability in the safety training \cite{ha2020SACLagLevine}.

\subsection{Action selection mechanism}
\label{subsection:Action selection mechanism}

\begin{algorithm}[H]
    \begin{algorithmic}[1]
    \STATE \textbf{Inputs:} $s, g, \SafetyActivated, \thGCtoS, \thStoGC$ 
    \STATE $\aGC \sim \piGC(.|s, g)$ ; $\aS \sim \piS(.|s)$
    \STATE $\text{lower} \leftarrow (\hat{\sigma}^{\piS}(s, \aGC) > \thStoGC)$
    \STATE $\text{raise} \leftarrow (\hat{\sigma}^{\piS}(s, \aS) > \thGCtoS)$
    \STATE $\SafetyActivated \leftarrow \SafetyActivated \, \text{\textbf{and}} \, \text{lower}$ \hfill // Lower the flag
    \STATE $\SafetyActivated \leftarrow \SafetyActivated \, \text{\textbf{or}} \, \text{raise}$  \hfill // Raise the flag
    \STATE $a \leftarrow \SafetyActivated . \aS + (1 - \SafetyActivated) . \aGC$
    \STATE \textbf{Return} $a$, $\SafetyActivated$
    \end{algorithmic}
    \caption{Action selection for safe exploration}
    \label{alg:Action selection}
\end{algorithm}

The action selection mechanism that must ensure safe exploration has been reproduced in \refAlg{alg:Action selection}.
It takes as input the current state $s$, current desired goal $g$, the current value of a boolean flag, called $\SafetyActivated$, and 
risk thresholds $\thGCtoS$ and $\thStoGC$. The flag indicates which action to choose, and the role of the action selection mechanism is to update this flag. If equal to $1$, action $\aS$ 
sampled by the safety policy is selected, else action $\aGC$ sampled by the GC policy is selected.
The flag is updated according to the level of risk associated  
with state $s$ and possible actions $\aGC$ and $\aS$. If the level of risk $\hat{\sigma}^{\piS}(s, \aS)$ 
associated with state $s$ and action $\aS$ exceeds threshold $\thGCtoS$, the safety flag is set to $1$ (\textit{raised})
so as to avoid a future potentially dangerous situation. On the contrary, if the level of risk $\hat{\sigma}^{\piS}(s, \aGC)$ 
associated with state $s$ and action $\aGC$ goes below threshold $\thStoGC$, the safety flag is set to $0$ (\textit{lowered}).
Note that the risk function $\hat{\sigma}^{\piS}$ is related to the safety policy, as its goal is to evaluate the capacity 
of the safety policy to put the agent out of danger. 
Also, because the safety policy is optimized on its critics, it tends to minimize the risk function undirectly. As a result, 
the thresholds should verify $\thStoGC \le \thGCtoS$. Otherwise, the safety policy would be the only one to act.
Depending on the environment, one could choose either equality or strict inequality between thresholds 
(See section \ref{subsec:ablation_dist}). 

The framework defined in section \refSection{subsection:safe_exp_problem} essentially leads to two 
definitions of the risk function, leading to three different strategies. 
The first is based on the time necessary to reach the set $\neighborhood$ from current state, using the safety policy. 
The second is based on the maximum of constraint function $h$ along future trajectories generated by the safety policy.
The third possibility is to combine both. We will use the labels \textbf{Time}, \textbf{Constraint} and \textbf{Time-constraint}
for the respective three strategies. 

\paragraph{\textbf{Time:}} We choose the set $\neighborhood$ so that, if the safety policy is well trained, 
then $\neighborhood$ is a forward set regarding the safety policy $\piS$. More precisely, we assume that, 
for any sequence of transitions $(s_0, a_0, ..., s_t, ...)$ generated by $\piS$, if 
the starting state $s_0$ lies in $\neighborhood$, then all future states $s_t$ also lie in $\neighborhood$.
Under this assumption the safety reward $\rS(s_t, a_t)$ equals $0$ until $s_t$ lies in $\neighborhood$, resulting in equation 
\refEq{eq:time_and_rewards}:

\begin{equation}
    T^{\piS}(s_0, a_0) = f_{\gamma}\left(\sum_{t = 0}^{+\infty} \gamma_t \rS(s_t, a_t)\right) 
    \label{eq:time_and_rewards}
\end{equation}
where $T^{\piS}(s_0, a_0)$ is the random variable corresponding to the number of steps necessary for $\piS$ to reach 
$\neighborhood$ starting from state-action couple $(s_0, a_0)$, and $f_{\gamma}(x) = \log \left( (1-\gamma)  x\right) / \log \gamma$,
which is a continuous bijection from $]0, 1/(1-\gamma)]$ to $\R^+$. By applying the bijective mapping $f_{\gamma}(x)$ to the atoms $\theta_{\psi_i}^{(j)}(s, a)$ 
of ensemble $Z_{\psi_1}, ... Z_{\psi_M}$, we obtain new atoms that approximate the distribution 
of the random variable $T^{\piS}(s, a)$.
We denote $\hat{T}^{\piS}(s, a ; \tau)$ the mean of the quantiles corresponding to cumulative probabilities 
greater than $\tau$. For example, $\tau = 0.9$ corresponds to the mean of worst $10\%$ of cases. $\tau$
is a hyperparameter of the algorithm. In the Time strategy : $\hat{\sigma}^{\piS}(s, a) = \hat{T}^{\piS}(s, a ; \tau)$.
So the thresholds $\thGCtoS$ and $\thStoGC$ are given in number of environment steps.

\paragraph{\textbf{Constraint:}}
In the same way, we denote $\hat{R}^{\piS}(s, a ; \tau)$ the mean of atoms from reachability critics $R_{\xi_1}, ... R_{\xi_M}$
corresponding to cumulative probabilities greater than $\tau$. 
In the constraint strategy : $\hat{\sigma}^{\piS}(s, a) = \hat{R}^{\piS}(s, a ; \tau)$.
So the thresholds $\thGCtoS$ and $\thStoGC$ are negative real numbers corresponding to safety margins.

\paragraph{\textbf{Time-constraint:}} This strategy combines the two previous ones. Let $\epsilon > 0$ be a positive real number 
representing a safety margin. If $\hat{R}^{\piS}(s, a ; \tau) > -\epsilon$, then: $\hat{\sigma}^{\piS}(s, a) = T_{max}$. 
Where $T_{max}$ is the maximum number of episode steps. 
Otherwise: $\hat{\sigma}^{\piS}(s, a) = \hat{T}^{\piS}(s, a ; \tau)$. 
The thresholds $\thGCtoS$ and $\thStoGC$ are given in number of environment steps.

\subsection{Algorithm}

\paragraph{\textbf{Pretraining phase:}} We use the same training loop as TQC 
\cite{TQC}. The only difference is the gradient step update where, in addition to the TQC update,
parameters $\xi_1, ... \xi_M$ are updated by performing Adam optimizer step on 
loss $\mathcal{L}_{Z}$ with target \refEq{eq:RCRL_target}, and the actor loss is replaced with loss $\mathcal{L}_{\pi_S}$.
\refAlg{alg:safety_update} describes how the update is done.
The safety policy interacts with the environment, generating a
transition that is stored in a replay buffer $\buffer_S$. Then a batch of transtions is sampled uniformly
from the buffer and a gradient step update is performed on all parameters, so that for each stored transition,
one step of gradient is performed. 
Also, we start the training after $5000$ steps for which we sample random actions
to favor exploration and learning at the start.

\begin{algorithm}[H]
\begin{algorithmic}[1]
\STATE Sample a batch from the replay buffer $\buffer_S$ uniformly
\STATE Perform TQC critic update on $\psi_1, ... \psi_M$ \cite{TQC}
\STATE Update $\xi_1, ... \xi_M$ using Adam on loss $\mathcal{L}_{Z}$ with target \refEq{eq:RCRL_target}
\STATE Update actor parameters $\phi_S$ using Adam on loss $\mathcal{L}_{\pi_S}$
\STATE Update temperature $\alpha_S$ according to SAC rule \cite{SAC}
\end{algorithmic}
\caption{Safety gradient step update}
\label{alg:safety_update}
\end{algorithm}

\paragraph{\textbf{Safe exploration phase:}} The procedure is summarized in \refAlg{alg:main_algorithm}, 
which is also off-policy. 
We first choose safety thresholds depending on the risk level we want. 
Then the parameters of the previously learned safety policy and its critics are loaded,
while the parameters related to the GC policy are randomly initialized. The episodic replay buffer is 
initially empty. For each environment step during training, the action selected according to 
\refAlg{alg:Action selection} to ensure safety during exploration. Each transition is stored in an 
episodic replay buffer $\buffer$ regardless of the policy that generated it. As a result, in a single stored episode,
there are probably samples generated by both policies. 

Initially, the GC policy performs poorly, as it has not yet been trained.
Thus the first stored episodes contain a large majority of transitions generated by the safety policy, as it had to
make up for the random behavior of the GC policy. Then, as the GC policy improves itself, the proportion of transitions 
it has generated increases in the buffer. The fact that many transitions have not been generated by the GC policy can 
lead off-policy algorithms like SAC to value over-estimation, due to the distributional drift between the dataset 
and the current learned policy \cite{levine2020offline}. This is why we used SAC-N instead of SAC, which is the 
same algorithm as SAC but with $N$ critics instead of $2$ \cite{sac_n_edac}. As the critics are initialized and updated
separately, disagreement between them regarding unvisited states is likely to lead the Bellman target to low values via the 
$\min$ operator, thus preventing over-estimation. 

\begin{algorithm}[H]
\begin{algorithmic}[1]
%\STATE \textbf{Initialize} Safety and GC parameters ;  %$\psi_1... \psi_M$, $\xi_1... \xi_M$, $\phi_S$, $\alpha_S$
%\STATE \textbf{Initialize} GC parameters %$\psi_{GC}... \psi_M$, $\xi_1... \xi_M$, $\phi_S$, $\alpha_S$
\STATE \textbf{Inputs:} $\thGCtoS, \thStoGC$ 
\STATE \textbf{Load} Safety parameters $\psi_i, \overline{\psi}_i$, $\xi_i, \overline{\xi}_i$, $i\in[\![1, M]\!]$, $\phi_S$, $\alpha_S$
\STATE \textbf{Initialize} GC parameters $\psi_{GC}, \overline{\psi}_{GC}$, $\phi_{GC}$, $\alpha_{GC}$
\STATE \textbf{Initialize} Episodic replay buffer $\buffer \leftarrow \emptyset$
\STATE \textbf{Sample} initial state $s_0 \sim p_0$, and goal $g \sim \pG$ 
\FOR{each iteration}
\FOR{each environment step, until done}
\STATE $a_t, \SafetyActivated \leftarrow \text{select\_action}(s, g, \SafetyActivated,$ \\$ \thGCtoS, \thStoGC)$
(\refAlg{alg:Action selection})
\STATE Collect transition $(s_t, a_t, s_{t+1}, r_{S, t}, r_{GC, t}, h_{t+1})$
\STATE $\buffer \leftarrow \buffer \cup \{ (s_t, a_t, s_{t+1}, r_{S, t}, r_{GC, t}, h_{t+1}) \}$
\ENDFOR 
\FOR{each gradient step}

\STATE Sample a batch $b_{GC} = (s_t, a_t, s_{t+1}, r_{GC, t})$ from $\buffer$ \\ using HER \cite{HER}
\STATE Update GC parameters on $b_{GC}$ using SAC-N rule \cite{SACN_edac}
% \IF{safety finetuning activated}
% \STATE Update safety policy using \refAlg{alg:safety_update}
% \ENDIF
\ENDFOR
\ENDFOR
\STATE \textbf{return} GC parameters $\psi_{GC}, \overline{\psi}_{GC}$, $\phi_{GC}$, $\alpha_{GC}$ %\\
%Safety parameters $\psi_i, \overline{\psi}_i$, $\xi_i, \overline{\xi}_i$, $i\in[\![1, M]\!]$, $\phi_S$, $\alpha_S$
%Replay buffer $\buffer$
\end{algorithmic}
\caption{Safe exploration algorithm}
\label{alg:main_algorithm}
\end{algorithm}

\section{Experiments}

To appreciate the performance of our method, we conducted experiments on two environments: a custom goal-based version of the gymnasium CartPole environment \cite{towers2024gymnasium} with continuous actions that we call \textit{CartPoleGC}, and a goal environment based on the Skydio X2 drone from the Mujoco menagerie \cite{menagerie2022github} that we call \textit{SkydioX2GC}. We performed an ablation study to validate our design choices and compared our approach with the state-of-the-art GC method SAC + HER. Finally, as our general objective is zero error during exploration, we analyze the causes of the few mistakes we obtained during safe exploration for the best strategy.

\subsection{Experimental setup}

CartPoleGC and SkydioX2GC are interesting environments due to their inherent instability as random actions or even actions near 0 lead to mistakes.

\begin{figure}[ht]
    \begin{subfigure}{0.25\textwidth}  % 0.3
        \centering 
        \includegraphics[width=\textwidth]{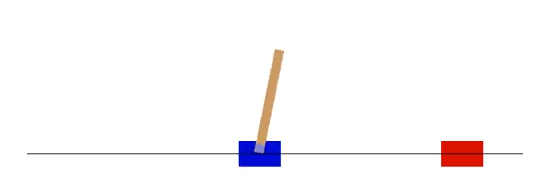}
        \caption{CartPoleGC with a goal in red}
        \label{fig:cartpole_gc}
    \end{subfigure}
    \hfill
    \begin{subfigure}{0.14\textwidth} % 0.16
        \centering
        \includegraphics[width=\textwidth]{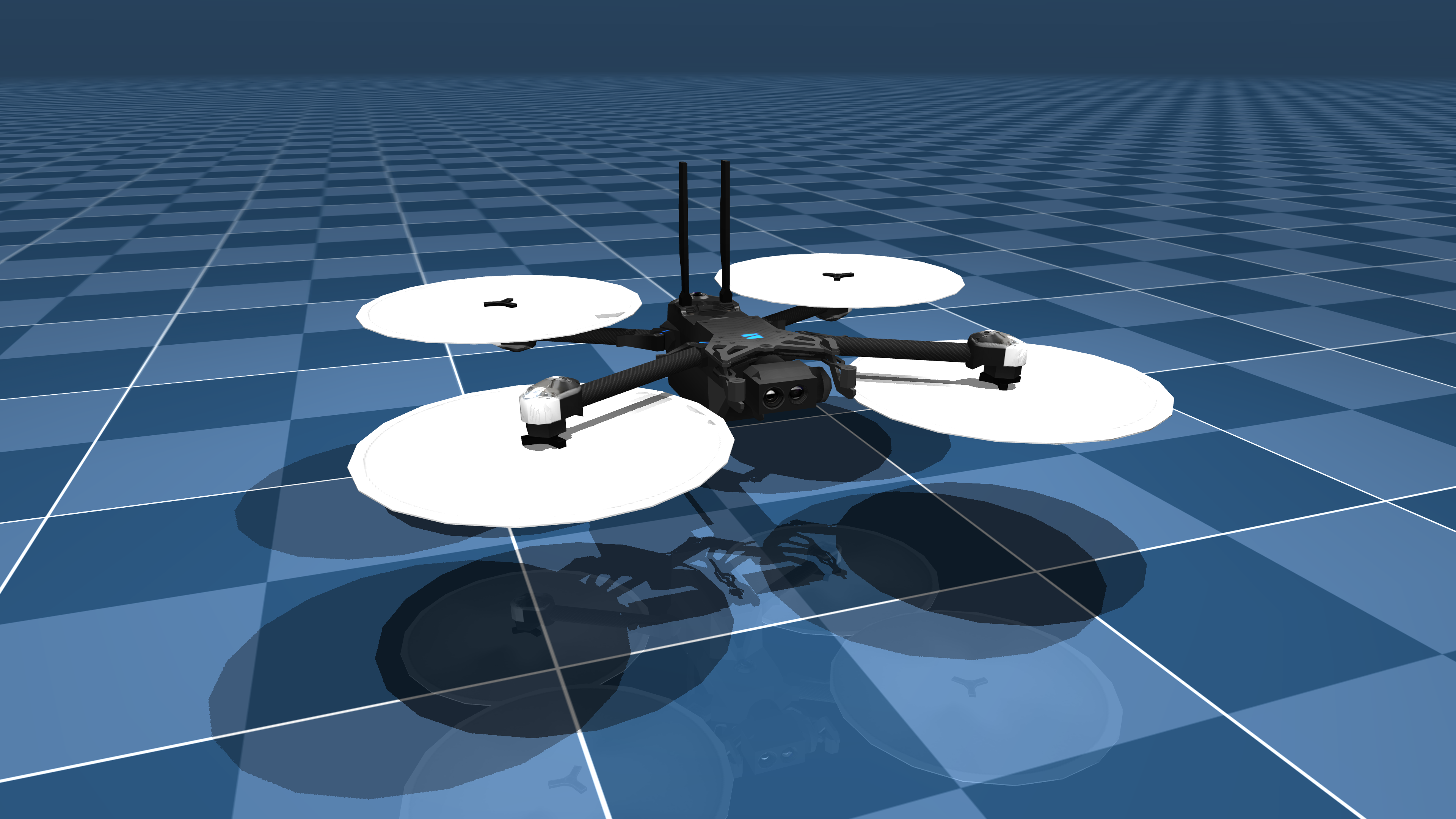}
        \caption{SkydioX2GC}
        \label{fig:x2}
    \end{subfigure}
    \caption{Environments}
    \Description{Environments}
    \label{fig:envs}
\end{figure}

% \begin{figure}[ht]
%     \centering
%     \includegraphics[width=0.4\textwidth]{images/cartpole_gc.png}
%     \includegraphics[width=0.1\textwidth]{images/x2.png}
%     \caption{CartpoleGC environment}
%     \Description{CartpoleGC environment}
%     \label{fig:cartpole_gc}
% \end{figure}

\subsubsection{CartPoleGC}
The state $s = (x, \dot{x}, \theta, \dot{\theta}) \in \stateSpace$ contains the position $x$ of the cart, its velocity $\dot{x}$, the angle of the pole $\theta$, and its 
velocity $\dot{\theta}$. $x$ lies in $[-2.4m, 2.4m]$, and $\theta$ in $[-0.41rad, 0.41rad]$.
Getting out of these bounds is considered a mistake, leading to episode termination. 
Action $a \in \actionSpace = [-1, 1]$ is proportional to the lateral force applied to the cart. 
$1$ corresponds to $1$ in the original discrete version and $-1$ to $0$. 
Goals are desired positions of the cart: $\goalSpace = [-2.16, 2.16]$, sampled uniformly. 
The agent gets a goal reward when the cart is less than $0.05m$ near the the desired goal.
The agent gets a safety reward when the $x$ position lies in $[-2.2, 2.2]$ and other state variables in 
$[-0.05, 0.05]$, so $\neighborhood = [-2.2, 2.2] \times [-0.05, 0.05]^3$.

\subsubsection{Skydio X2}
The system physics is fully characterized by position $(x, y, z)$ of the center of gravity, orientation (quaternions),
linear velocity $(\dot{x}, \dot{y}, \dot{z})$ and angular velocity (Euler angles).
The state is the concatenation of these elements. The $(x, y)$ position lies in $[-3m, 3m]^2$, altitude $z$ in $(1m, 4m)$, 
while roll and pitch angles are bounded to $30 \deg$ of amplitude around $0$. 
Getting out of these bounds is considered a mistake, leading to episode termination. 
The action space $\actionSpace = [-1, 1]^4$ corresponds to the forces applied on the rotors.
Goals are desired $(x, y, z)$ positions in the cube $[-2.6m, 2.6m] \times [-2.6m, 2.6m] \times [1.4m, 3.6m]$,
sampled uniformly. The agent gets a goal reward when its less than $0.25m$ near the goal, and a safety reward when
it is in the cube $\neighborhood = [-2.8m, 2.8m] \times [-2.8m, 2.8m] \times [1.2m, 3.8m]$.

Function $h$ has the same structure in both environments. For a bounded state variable, \textit{eg} $x$ for CartPoleGC,
corresponds a function $h_x$, computed according to equation \refEq{eq:h_cartpole}. 
\begin{equation}
    h_x(x) = \max \{ -1 - 2 \frac{x - x_{middle}}{x_{max} - x_{min}}, 2 \frac{x - x_{middle}}{x_{max} - x_{min}} - 1 \}
    \label{eq:h_cartpole} 
\end{equation}
where $x_{middle} = (x_{max} + x_{min}) / 2$. The constraint value $h(s)$ is the maximum over all bounded state variables.
The advantage of such a formulation is that $h_x$ is bounded in $[-1.0, 0.]$, so that $\lambda \overline{R}^{\piS}$ and 
$\overline{Q}^{\piS}$ have the same scale (Cf \refSection{subsection:safety_policy_learning}). 
Finally, for both environments, an episode is terminated when $h(s) > 0$ and truncated when the number of steps
exceeds $500$. Note that achieving a goal is not terminal. Thus the GC policy can be aggressive to
reach the goal and maintain safety once the goal is reached. For both environments, reset anywhere is performed in pretraining, while in safe exploration a noisy reset centered on a single state is applied.

\subsubsection{Baseline} 
For each environment, we compared our best safe exploration variant in terms of safety with 
the combination of SAC and HER (SAC + HER), with $80\%$ of relabelling and 
\textit{future} strategy, 
in terms of success rate and occurrence of mistakes \cite{SAC, HER}.

\subsubsection{Comparisons settings}
We trained safe exploration variants and the baseline 12 times with CartPoleGC and 9 times with SkydioX2GC. We performed fewer runs on SkydioX2GC because of the computational cost. To obtain the 12 runs on CartPoleGC, we performed 4 pretrainings using 4 different seeds. Then, for each of these pre-trained safety policies, we trained 3 safe exploring agents with 3 seeds. The idea behind this is to avoid cherry picking and draw a fair comparison between algorithms as the quality of safe exploration strongly depends on the the pretraining. In the same way, we performed 3 pretrainings on Skydio with 3 different seeds and 3 safe explorations for each pretraining seed. Each pretraining takes $2$ million steps for CartPoleGC and $2.5$ million steps for SkydioX2GC. For safe exploration, it takes respectively $500\,000$ and $1$ million steps. 
We are interested in minimizing the occurrence of mistakes 
for the worst possible run of each tested variant. So we show the mean, the minimum, and the maximum in our plots of mistake occurrence. As for the success rate, we show the mean and the standard deviation. In all our 
experiments we set $\tau = 0.9$, considering the $10\%$ worst cases to compute the risk measure, and 
a margin $\epsilon = 0.1$ (Cf section \ref{subsection:Action selection mechanism}).

\subsection{Comparison with the baseline}

\begin{figure}[ht]
    \begin{subfigure}{0.22\textwidth}  
        \centering 
        \includegraphics[width=\textwidth]{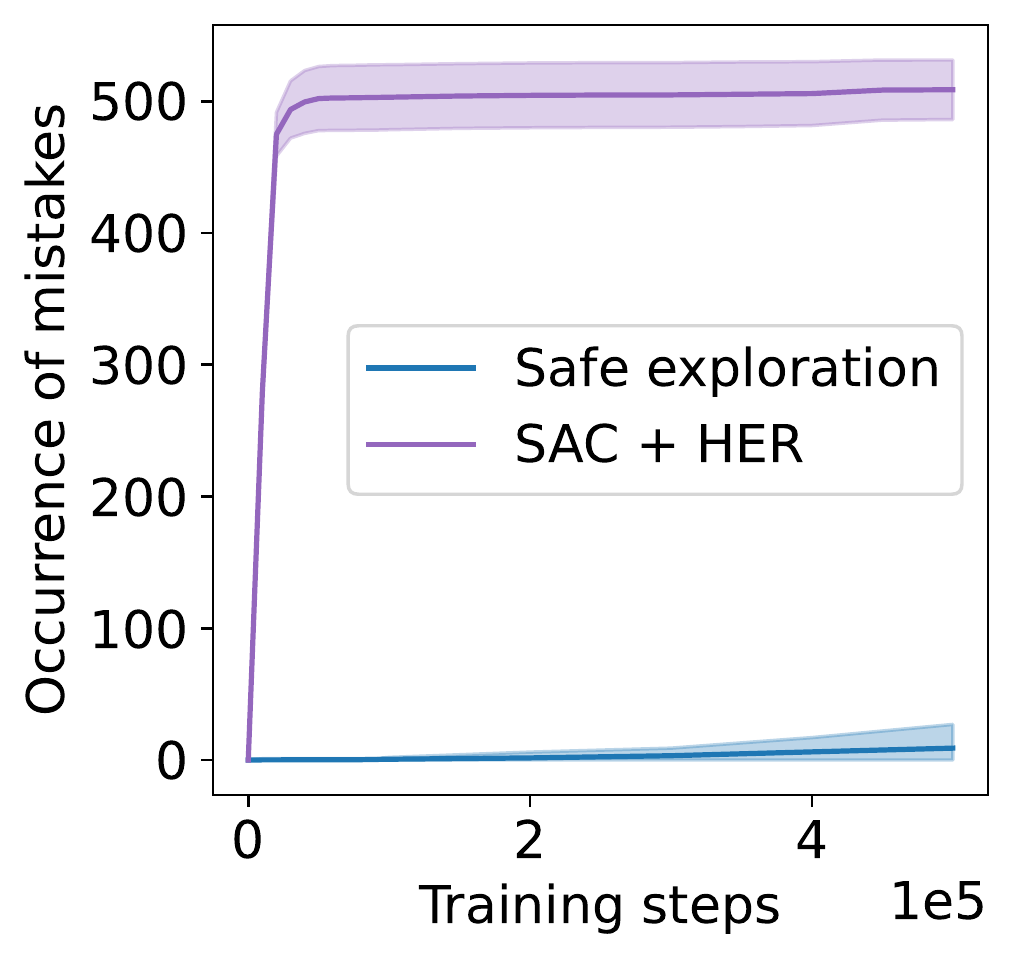}
        \caption{}
        \label{fig:cartpole_comp_baseline/terminated_regret}
    \end{subfigure}
    \hfill
    \begin{subfigure}{0.22\textwidth}
        \centering
        \includegraphics[width=\textwidth]{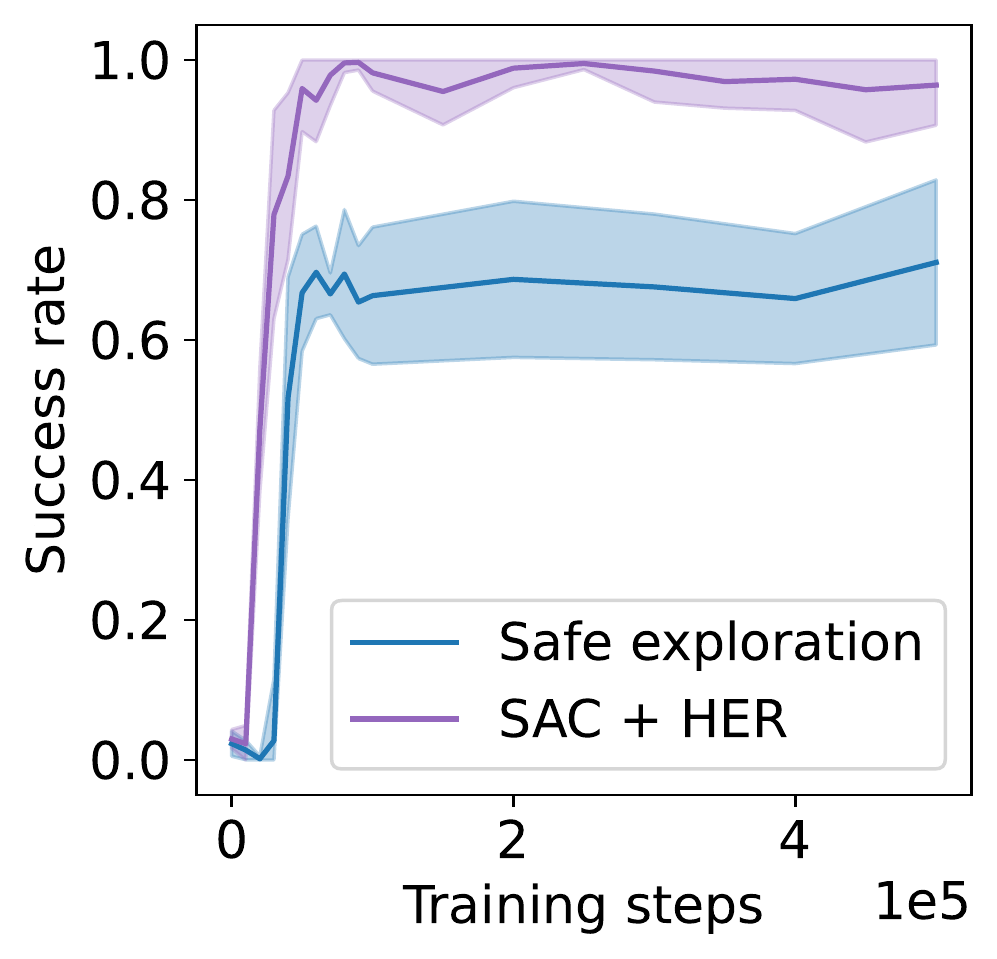}
        \caption{}
        \label{fig:cartpole_comp_baseline/is_success_final}
    \end{subfigure}
    \caption{Comparison between our method (Cf $L\&S$ in \refSection{subsec:ablation_dist}) and the baseline on the CartPoleGC environment in terms of safety during exploration and coverage.}
    \Description{Comparison between our method (Cf $L\&S$ in \refSection{subsec:ablation_dist})and the baseline on the CartPoleGC environment in terms of safety during exploration and coverage.}
    \label{fig:cartpole_comp_baseline}
\end{figure}

\begin{figure}[ht]
    \begin{subfigure}{0.22\textwidth}  
        \centering 
        \includegraphics[width=\textwidth]{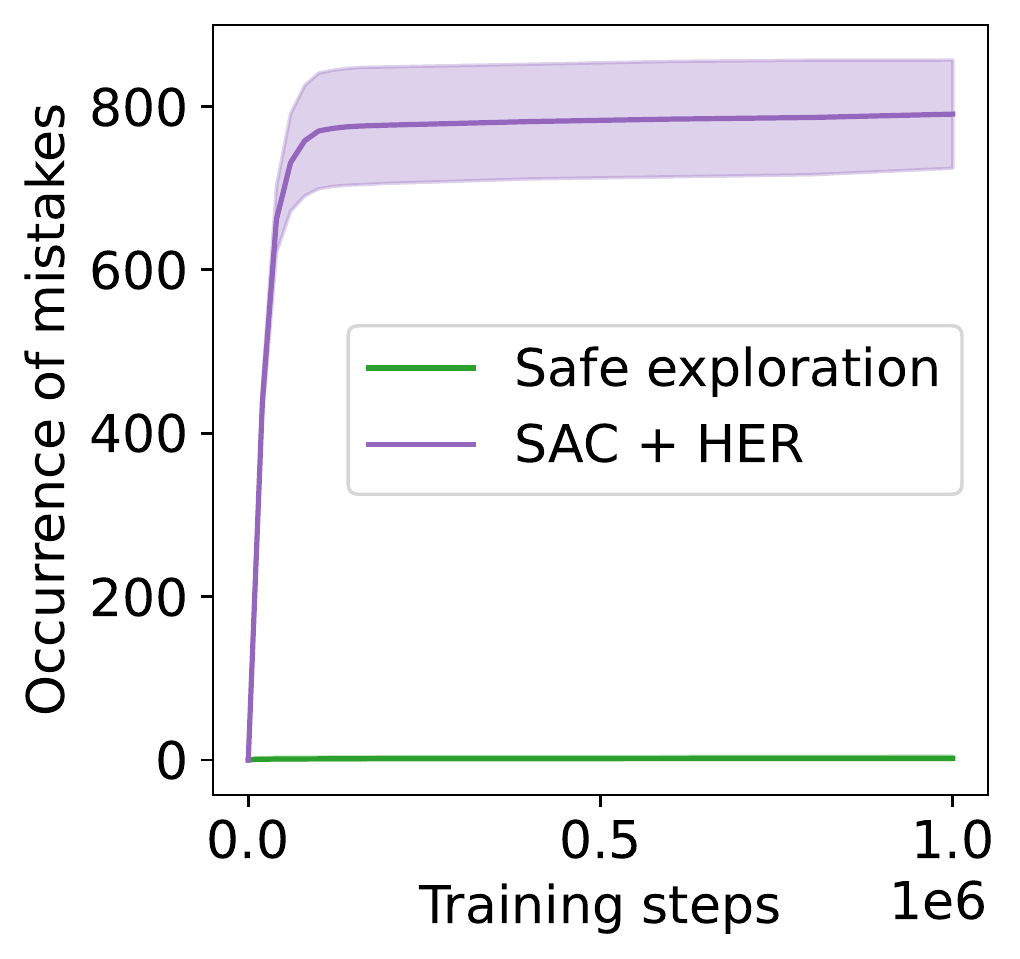}
        \caption{}
        \label{fig:skydio_comp_gc_vs_best_safe_exp/terminated_regret}
    \end{subfigure}
    \hfill
    \begin{subfigure}{0.22\textwidth}
        \centering
        \includegraphics[width=\textwidth]{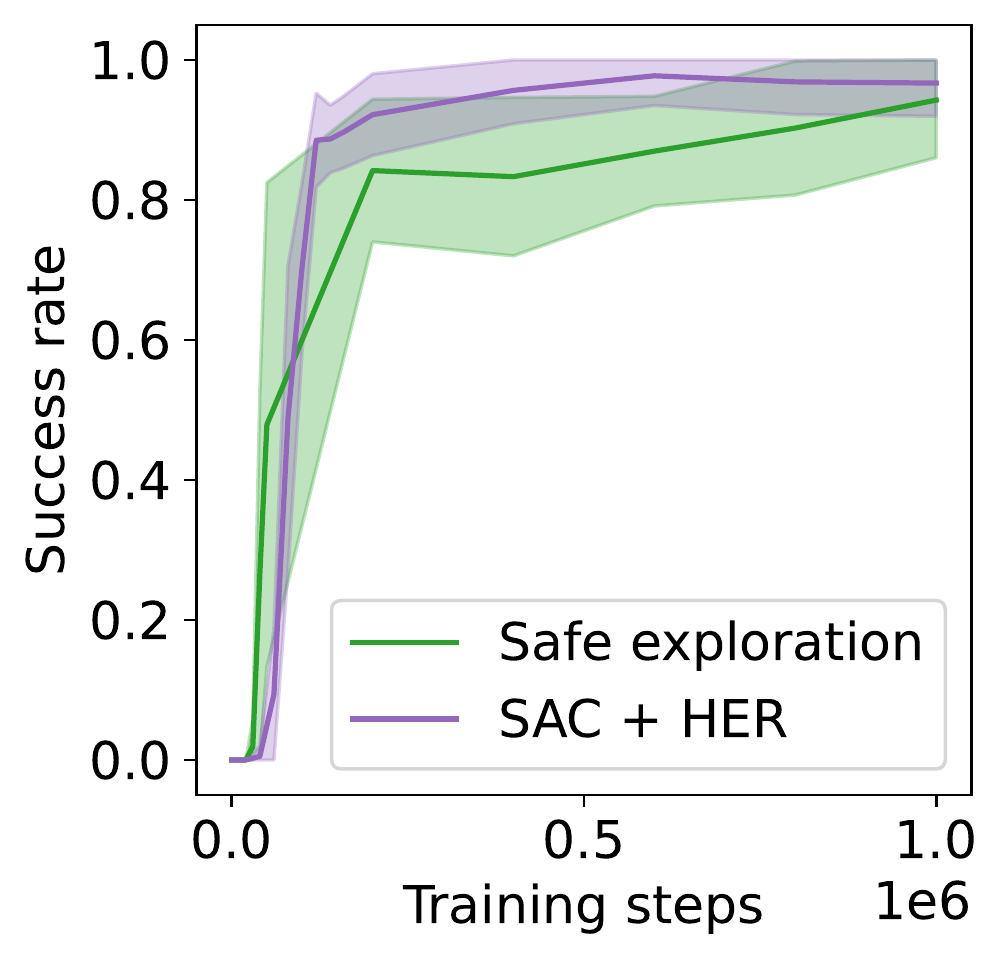}
        \caption{}
        \label{fig:skydio_comp_gc_vs_best_safe_exp/is_success_final}
    \end{subfigure}
    \caption{Comparison between our method (Cf $S$ in \refSection{subsec:ablation_dist}) and the baseline on the SkydioX2GC environment in terms of safety during exploration and coverage.}
    \Description{Comparison between our method (Cf $S$ in \refSection{subsec:ablation_dist}) and the baseline on the SkydioX2GC environment in terms of safety during exploration and coverage.}
    \label{fig:skydio_comp_gc_vs_best_safe_exp}
\end{figure}

For both environments, we show that our approach can considerably reduce the number of mistakes 
during exploration in comparison to the baseline, as it does not take safety into account
(Figures \ref{fig:cartpole_comp_baseline} and \ref{fig:skydio_comp_gc_vs_best_safe_exp}). 
However, the baseline obtains a better success rate, around $98\%$ on average against $70\%$ 
for CartPoleGC (\refFig{fig:cartpole_comp_baseline/is_success_final}) 
while on SkydioX2GC safe exploration offers almost the same performance in coverage but with higher
variance (\refFig{fig:skydio_comp_gc_vs_best_safe_exp/is_success_final}). 
CartPoleGC's poorer coverage can be explained by the fact that some goals are near terminal states
so the policy prevents the cart from reaching it. We leave a coverage map in the supplementary material.
Most importantly, we want to stress the minimum and maximum performance regarding safety.
With CartPoleGC we obtained at most 27 mistakes, and with 7 mistakes SkydioX2GC, but for some runs,
we obtained 0 mistakes for the whole training.
As shown in \refSection{subsection:analysing failures}, these differences 
correspond to different pre-trained safety policies.

\subsection{Ablation study}

\subsubsection{Effect of the reachability critics}
\label{subsec:ablation_dist}
In section \ref{subsection:Action selection mechanism}, we describe three strategies to compute 
the risk function: Time, Constraint, and Time-constraint, that we would like to rate in terms of safety during exploration and coverage. Also, the $\lambda$ factor used in pretraining can be set to $0$ or
$100$ to take reachability critics into account or not in the actor update. 
The constraint (only) strategy led to catastrophic results 
in our preliminary experiments, both for learning and action selection, 
so we left it in the supplementary material. Four classes remain: $L\&S$ (reachability is taken into account both in safety learning and action selection), $L$ (learning only), $S$ (action selection only), $None$ (reachability is not taken into account).

\begin{figure}[ht]
    \begin{subfigure}{0.22\textwidth}  
        \centering 
        \includegraphics[width=\textwidth]{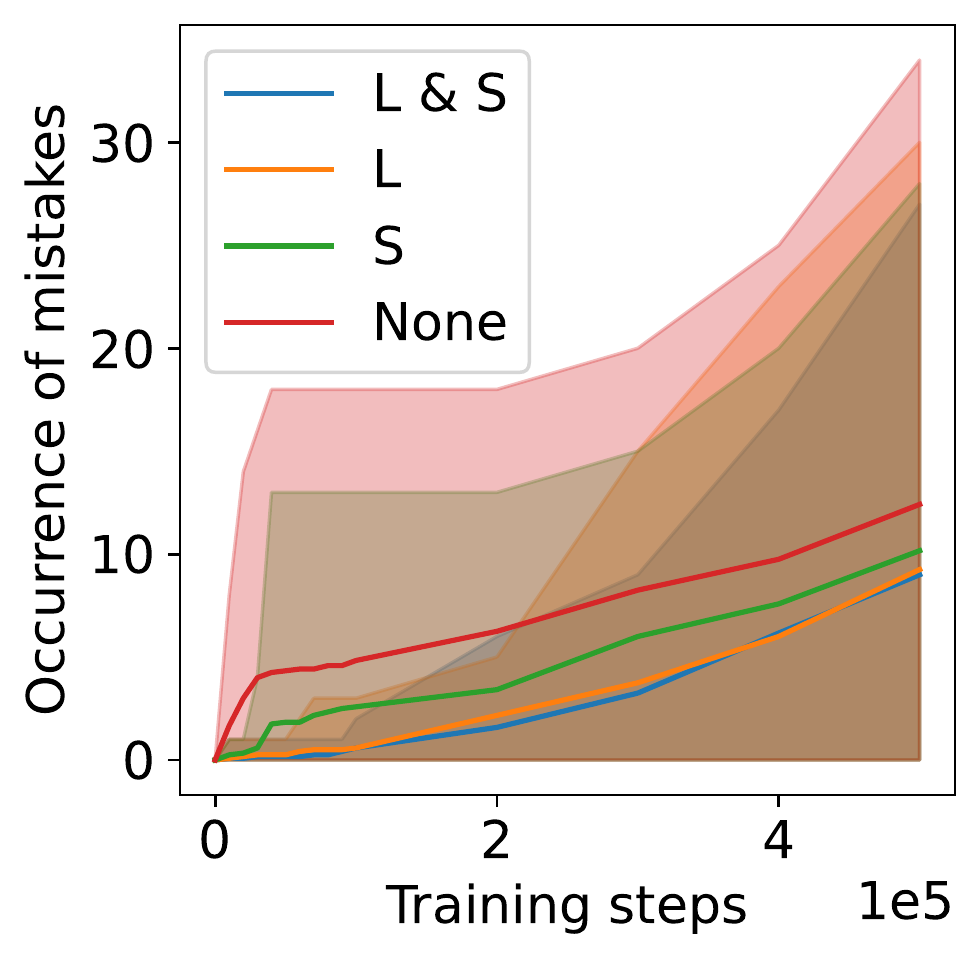}
        \caption{}
        \label{fig:cartpole_dist_ablation_full/terminated_regret}
    \end{subfigure}
    \hfill
    \begin{subfigure}{0.22\textwidth}
        \centering
        \includegraphics[width=\textwidth]{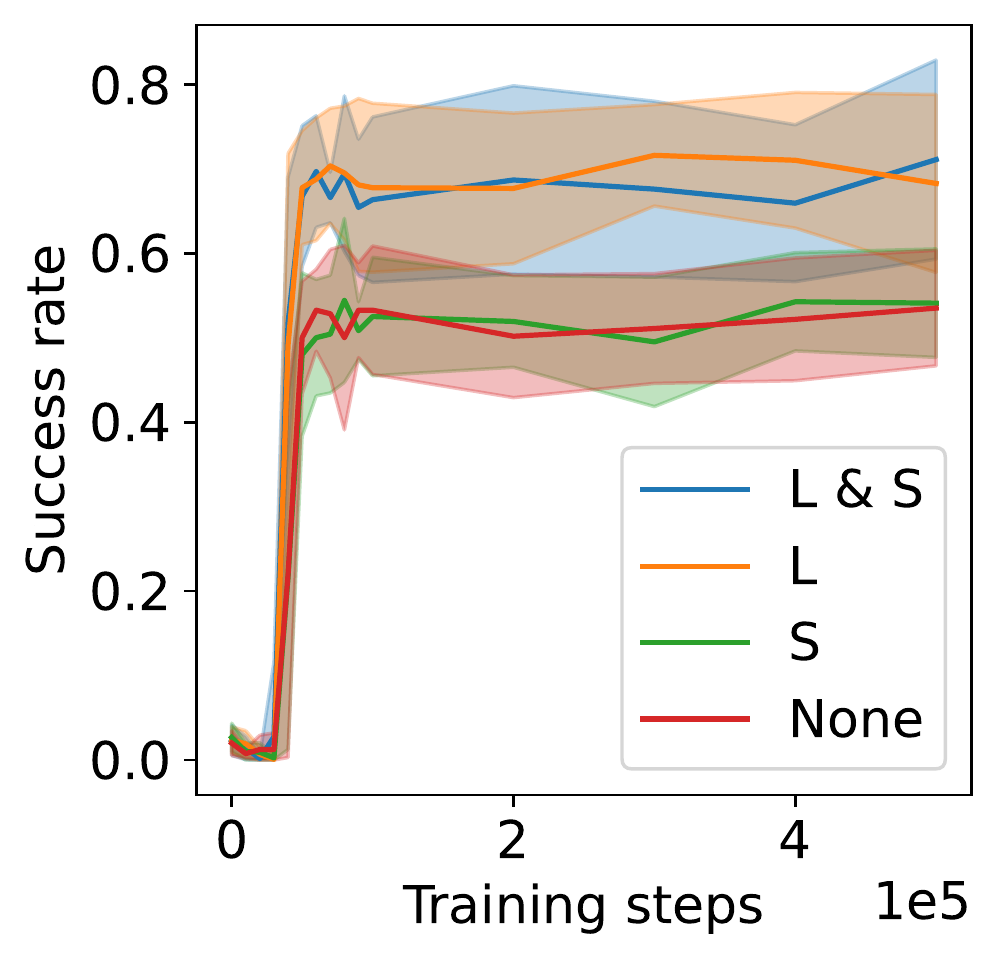}
        \caption{}
        \label{fig:cartpole_dist_ablation_full/is_success_final}
    \end{subfigure}
    \caption{Effect of the reachability critics on safe exploration with CartPoleGC for different variants. $L\&S$ (reachability in safety learning and action selection), $L$ (learning only), $S$ (action selection only), $None$ (no reachability)}
    \Description{Effect of the reachability critics on safe exploration with CartPoleGC for different variants. $L\&S$ (reachability in safety learning and action selection), $L$ (learning only), $S$ (action selection only), $None$ (no reachability)}
    \label{fig:cartpole_dist_ablation_full}
\end{figure}

\begin{figure}[ht]
    \begin{subfigure}{0.22\textwidth}  
        \centering 
        \includegraphics[width=\textwidth]{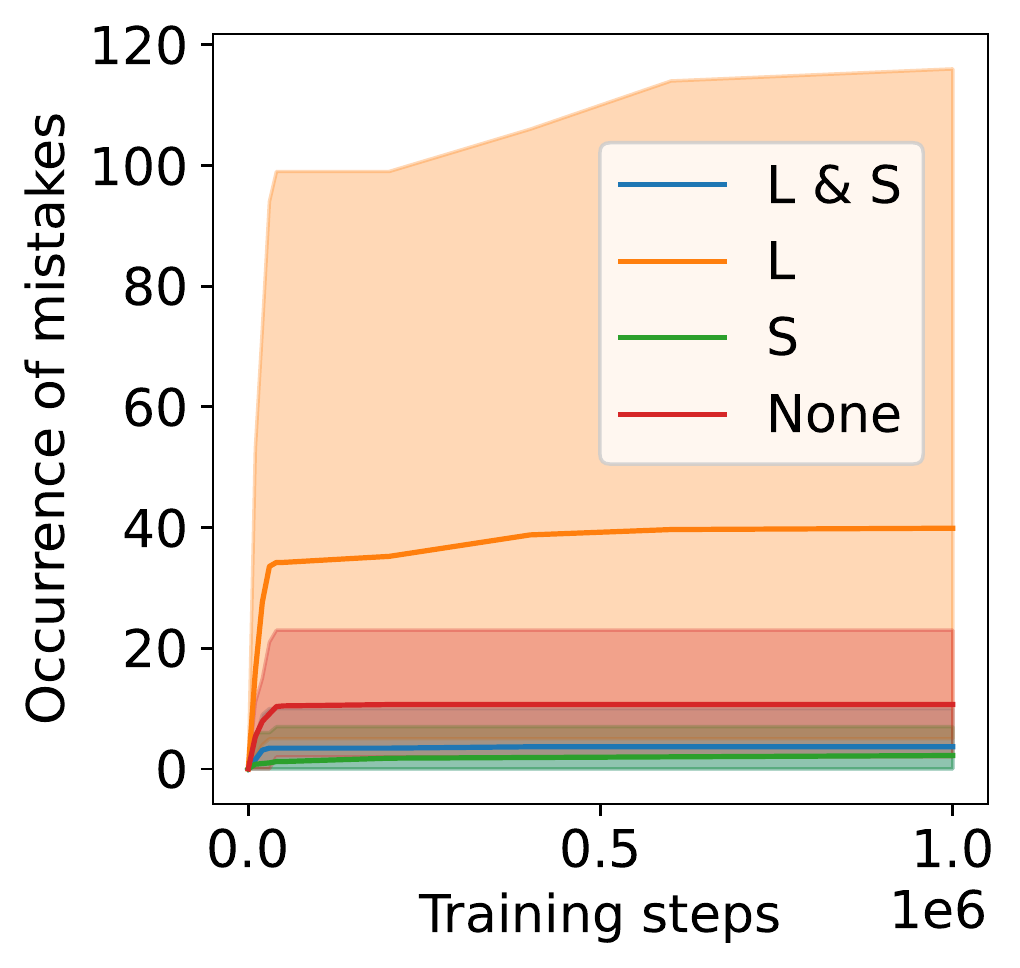}
        \caption{}
        \label{fig:skydio_ablation_dist_full_std/terminated_regret}
    \end{subfigure}
    \hfill
    \begin{subfigure}{0.22\textwidth}
        \centering
        \includegraphics[width=\textwidth]{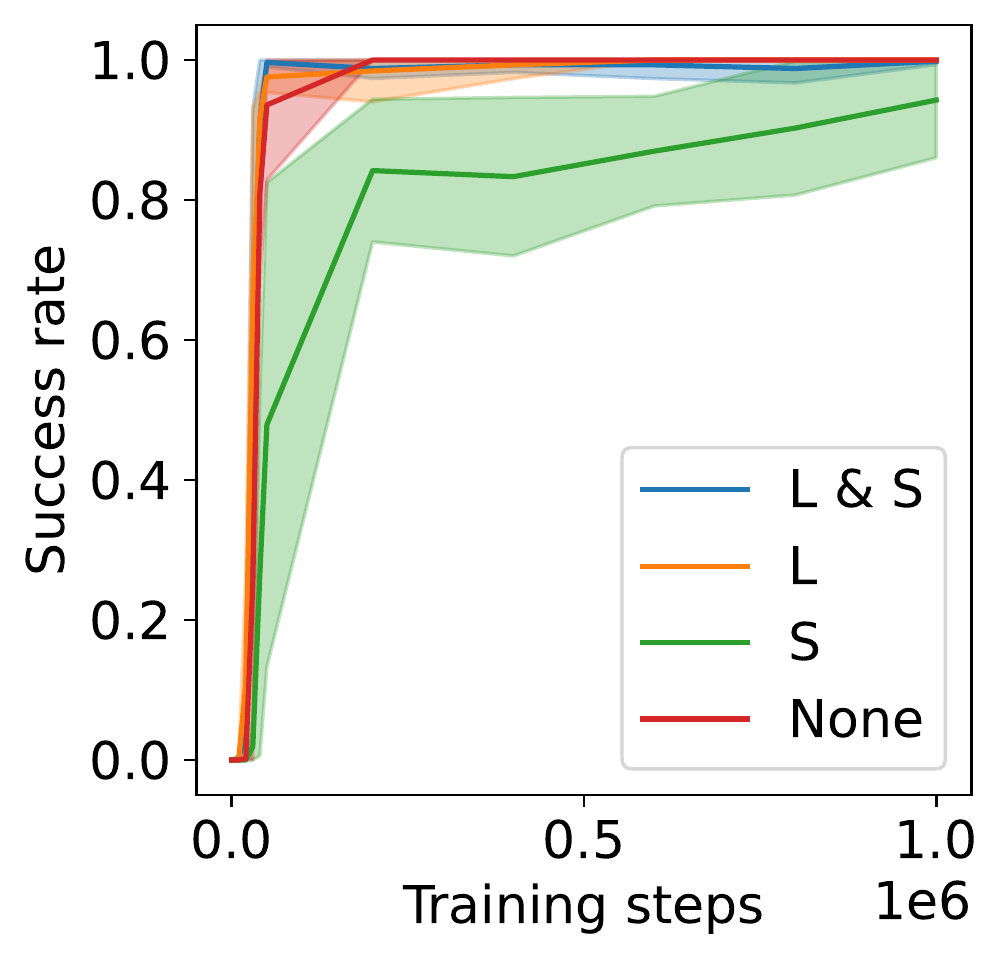}
        \caption{}
        \label{fig:skydio_ablation_dist_full_std/is_success_final}
    \end{subfigure}
    \caption{Effect of the reachability critics on safe exploration with SkydioX2GC for different variants. $L\&S$ (reachability in safety learning and action selection), $L$ (learning only), $S$ (action selection only), $None$ (no reachability)}
    \Description{Effect of the reachability critics on safe exploration with SkydioX2GC for different variants. $L\&S$ (reachability in safety learning and action selection), $L$ (learning only), $S$ (action selection only), $None$ (no reachability)}
    \label{fig:skydio_ablation_dist_full_std}
\end{figure}

First, it is interesting to note the difference between the two environments. In the case of CartPoleGC, 
the $L\&S$ combination is the best as it obtains the least mistakes (\refFig{fig:cartpole_dist_ablation_full/terminated_regret}) 
with the highest coverage (\refFig{fig:cartpole_dist_ablation_full/is_success_final}). $S$ obtains the least 
number of mistakes for SkydioX2GC (\refFig{fig:skydio_ablation_dist_full_std/terminated_regret}) 
while it performs poorly on CartPoleGC both in safety and coverage \refFig{fig:cartpole_dist_ablation_full}.
Also, while $L$ offers the second-best safety performance for CartPoleGC, it obtains the worst safety performance 
with SkydioX2GC. Eventually, as $L\&S$ obtains a maximum of 10 mistakes against 7 for $S$ and clearly outperforms 
$S$ in terms of coverage, it appears as the default choice for both environments.
Also one may observe a common pattern in Figures \ref{fig:cartpole_dist_ablation_full} and 
\ref{fig:skydio_ablation_dist_full_std}. In both cases including the reachability critic in the actor loss
($L$ and $L\&S$) leads to better coverage than other approaches.
In the case of SkydioX2GC, we can see that other ablations than $S$ perform way better in terms of coverage.

\subsubsection{Effect of the thresholds and hysteresis}

\begin{figure}[ht]
    \begin{subfigure}{0.22\textwidth}  
        \centering 
        \includegraphics[width=\textwidth]{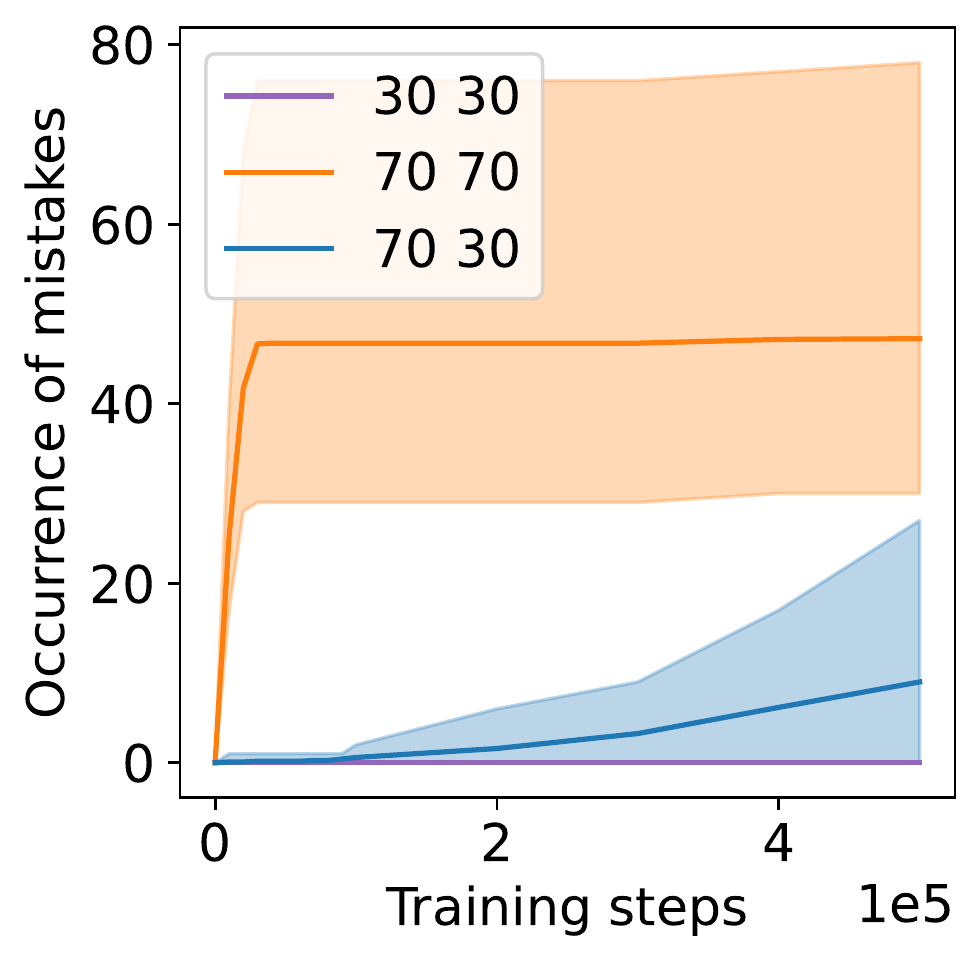}
        \caption{}
        \label{fig:cartpole_hysteresis/terminated_regret}
    \end{subfigure}
    \hfill
    \begin{subfigure}{0.22\textwidth}
        \centering
        \includegraphics[width=\textwidth]{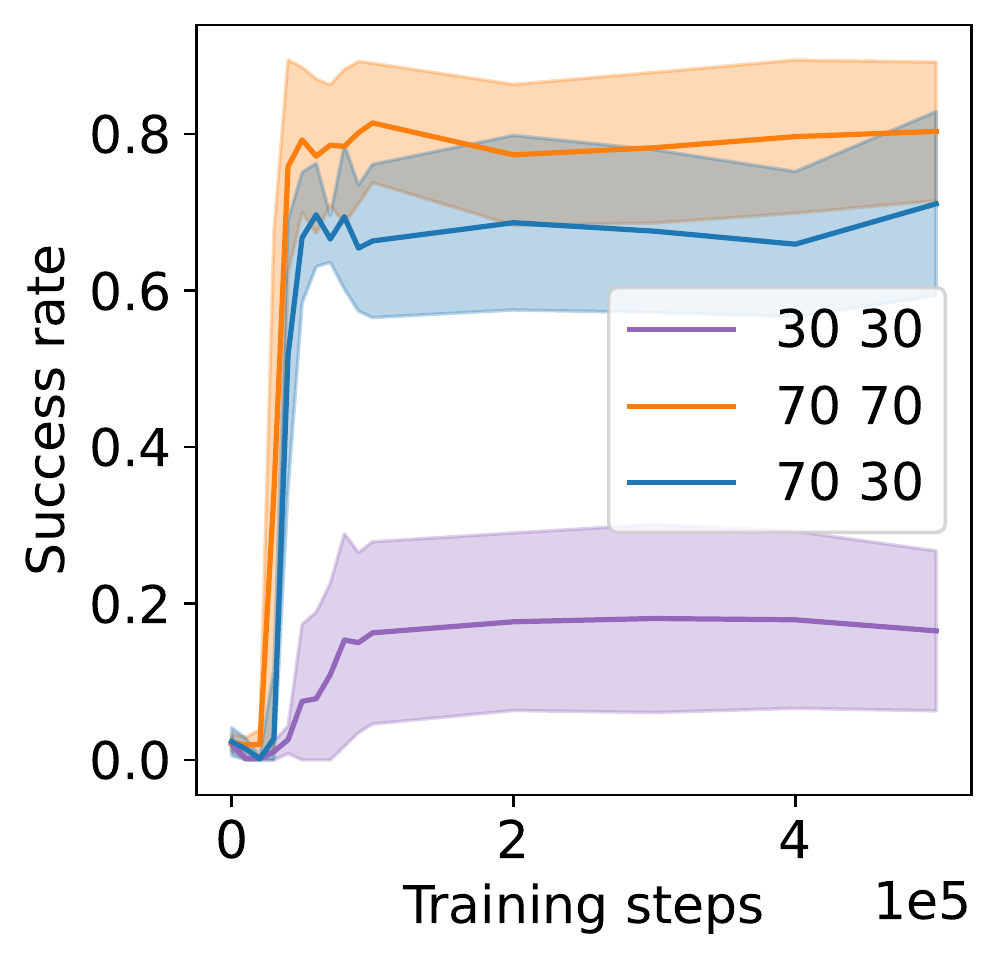}
        \caption{}
        \label{fig:cartpole_hysteresis/is_success_final}
    \end{subfigure}
    \caption{Effect of the thresholds on safe exploration with CartPoleGC for $(\thGCtoS, \thStoGC) \in \{ (30, 30), (70, 70), (70, 30)\}$ and the $L\&S$ variant.}
    \Description{Effect of the thresholds on safe exploration with CartPoleGC for $(\thGCtoS, \thStoGC) \in \{ (30, 30), (70, 70), (70, 30)\}$ and the $L\&S$ variant.}
    \label{fig:cartpole_hysteresis}
\end{figure}

By construction, our method implies a tradeoff between safety during exploration and coverage.
For instance, we remark that switching between the safety policy and the GC policy delays 
goal achievement in comparison to the full GC baseline. We leave videos in the supplementary material.
Again, the results are very different between environments. As for CartPoleGC, we show that high thresholds
(70 70 in \refFig{fig:cartpole_hysteresis})
lead to catastrophic performance in terms of safety during exploration, but also lead to
better coverage. On the contrary low thresholds (30 30 in \refFig{fig:cartpole_hysteresis})
lead to zero safety violations for all seeds but also to very poor coverage.
The best tradeoff between safety and coverage is obtained with $(\thGCtoS, \thStoGC) = (70, 30)$,
leading to a hysteresic behavior of the action selection mechanism. This choice was motivated by the 
fact that we have less confidence in the GC policy than in the safety policy to ensure safe exploration.
As for SkydioX2GC, the result is the exact opposite. Equal thresholds lead the the best safety performance 
while the hysteresic choice seems to generate way more instability.

\begin{figure}[ht]
    \begin{subfigure}{0.22\textwidth}  
        \centering 
        \includegraphics[width=\textwidth]{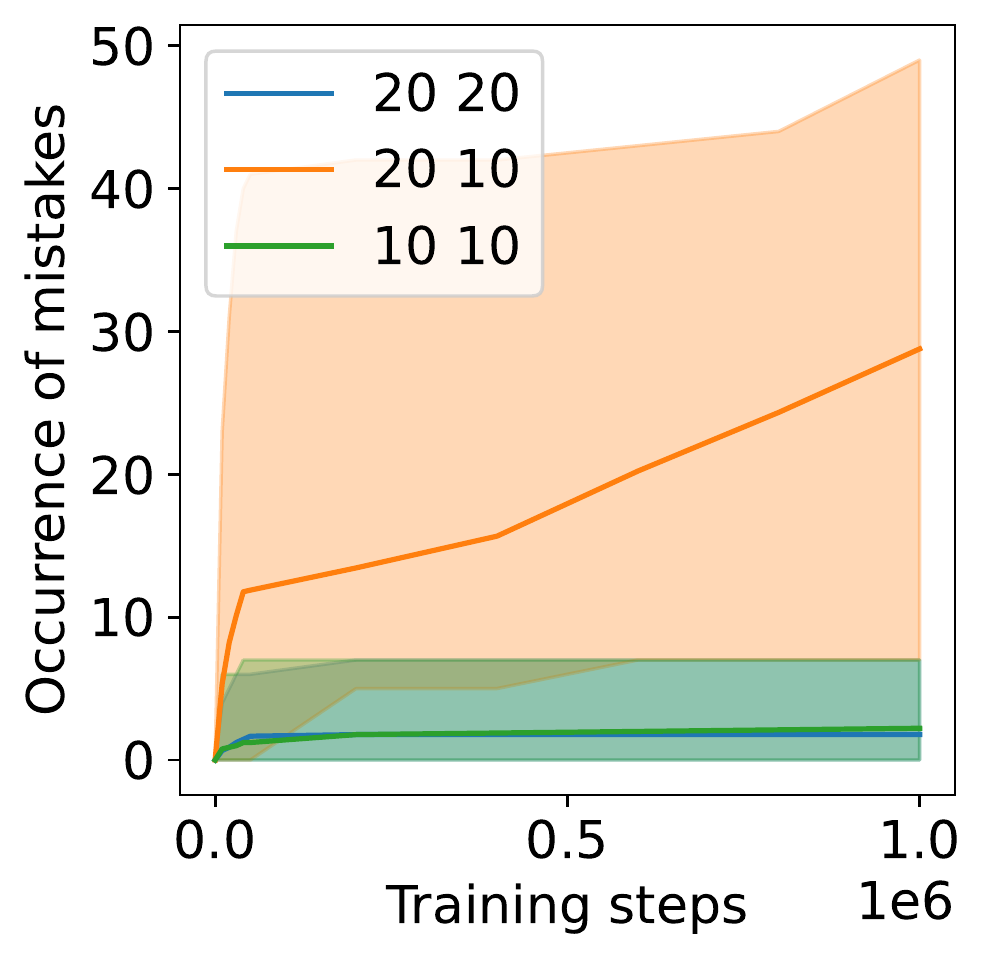}
        \caption{}
        \label{fig:skydio_hysteresis/terminated_regret}
    \end{subfigure}
    \hfill
    \begin{subfigure}{0.22\textwidth}
        \centering
        \includegraphics[width=\textwidth]{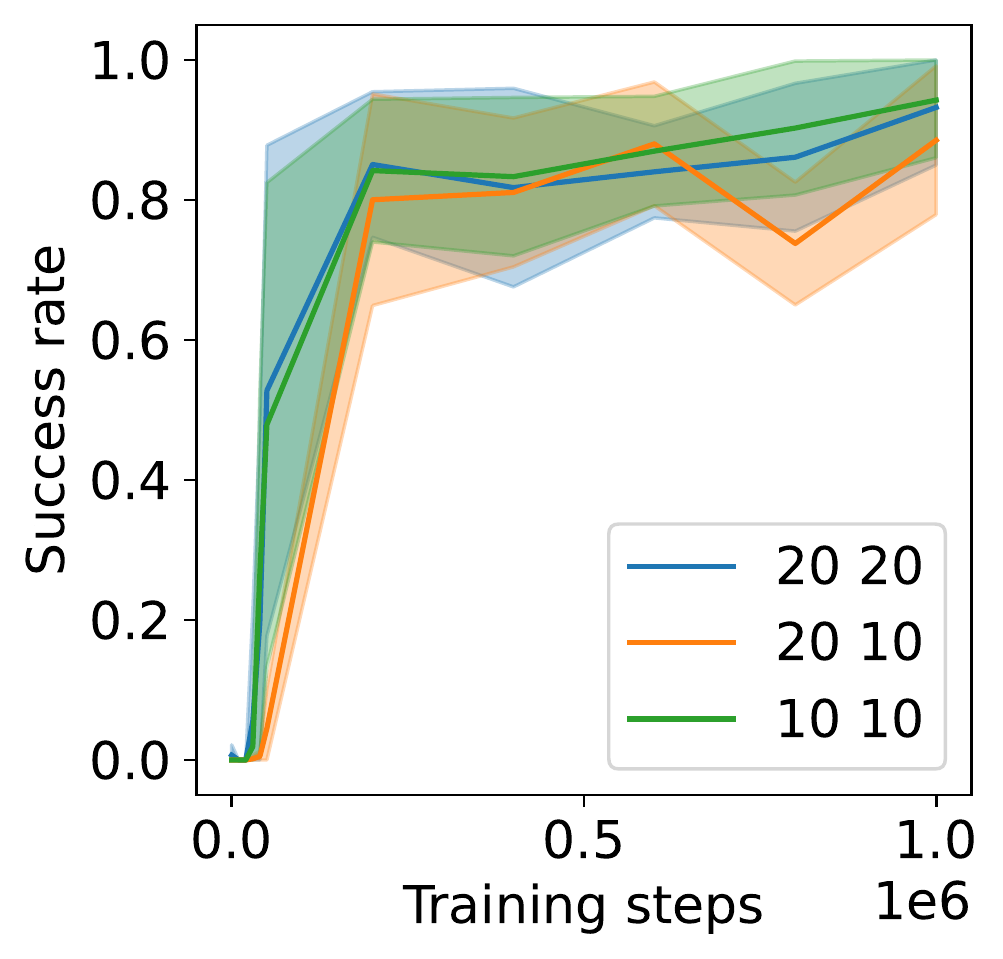}
        \caption{}
        \label{fig:skydio_hysteresis/is_success_final}
    \end{subfigure}
    \caption{Effect of the thresholds on safe exploration with SkydioX2GC for $(\thGCtoS, \thStoGC) \in \{ (10, 10), (20, 20), (20, 10)\}$ and the $S$ variant.}
    \Description{Effect of the thresholds on safe exploration with SkydioX2GC for $(\thGCtoS, \thStoGC) \in \{ (30, 30), (70, 70), (70, 30)\}$ and the $S$ variant.}
\end{figure}

\subsection{Analysing failure modes}
\label{subsection:analysing failures}

\begin{SCfigure}
\includegraphics[width=0.22\textwidth]{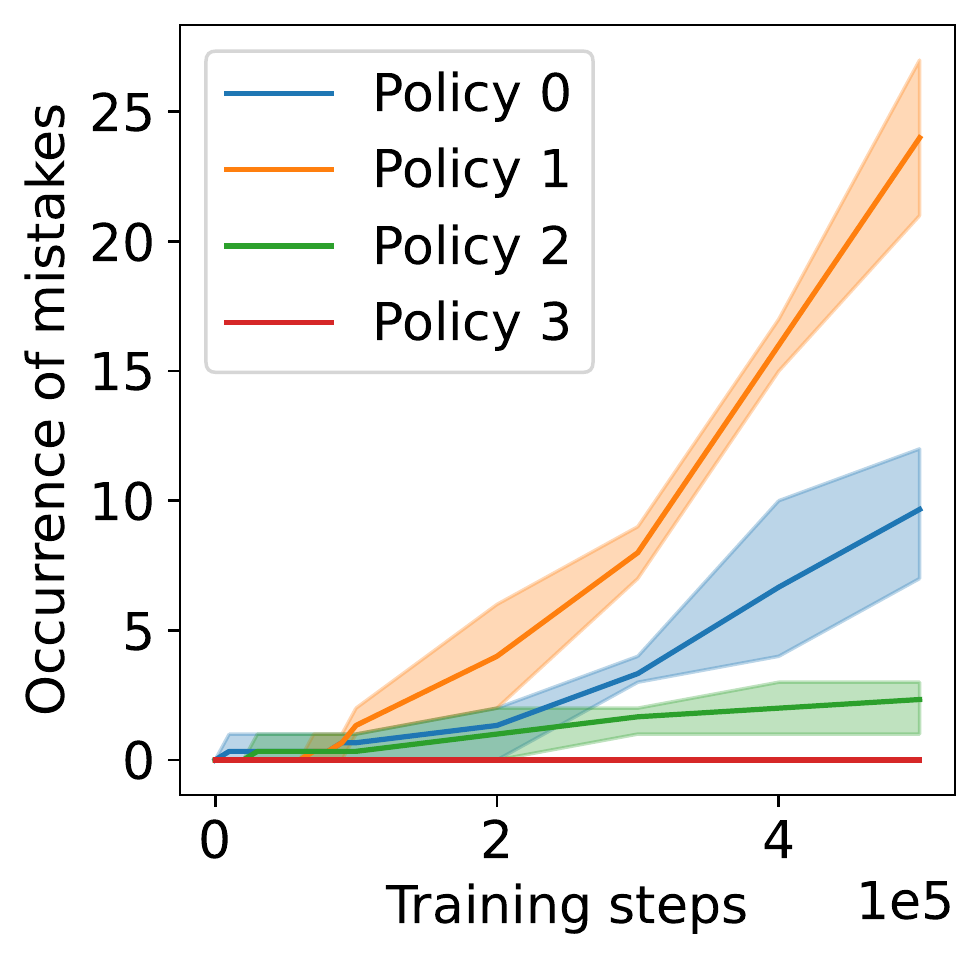}
\caption{Occurrence of mistakes obtained during safe exploration for different pre-train policies with the CartPoleGC environment and $(\thGCtoS, \thStoGC) = (70, 30)$}
\Description{Occurrence of mistakes obtained during safe exploration for different pre-train policies 
with the CartPoleGC environment and $(\thGCtoS, \thStoGC) = (70, 30)$}
\label{fig:pretrain_effect}
\end{SCfigure}

%\begin{figure}[ht]
%\centering
%\includegraphics[width=0.22\textwidth]{images/cartpole_different_learn_seed_70_30/terminated_regret.pdf}
%\caption{$}
%\Description{Occurrence of mistakes obtained during safe exploration for different pretraining policies 
%with the CartPoleGC environment, $(\thGCtoS, \thStoGC) = (70, 30)$}
%\label{fig:pretrain_effect}
%\end{figure}

\begin{figure}[ht]
    \includegraphics[width=0.4\textwidth]{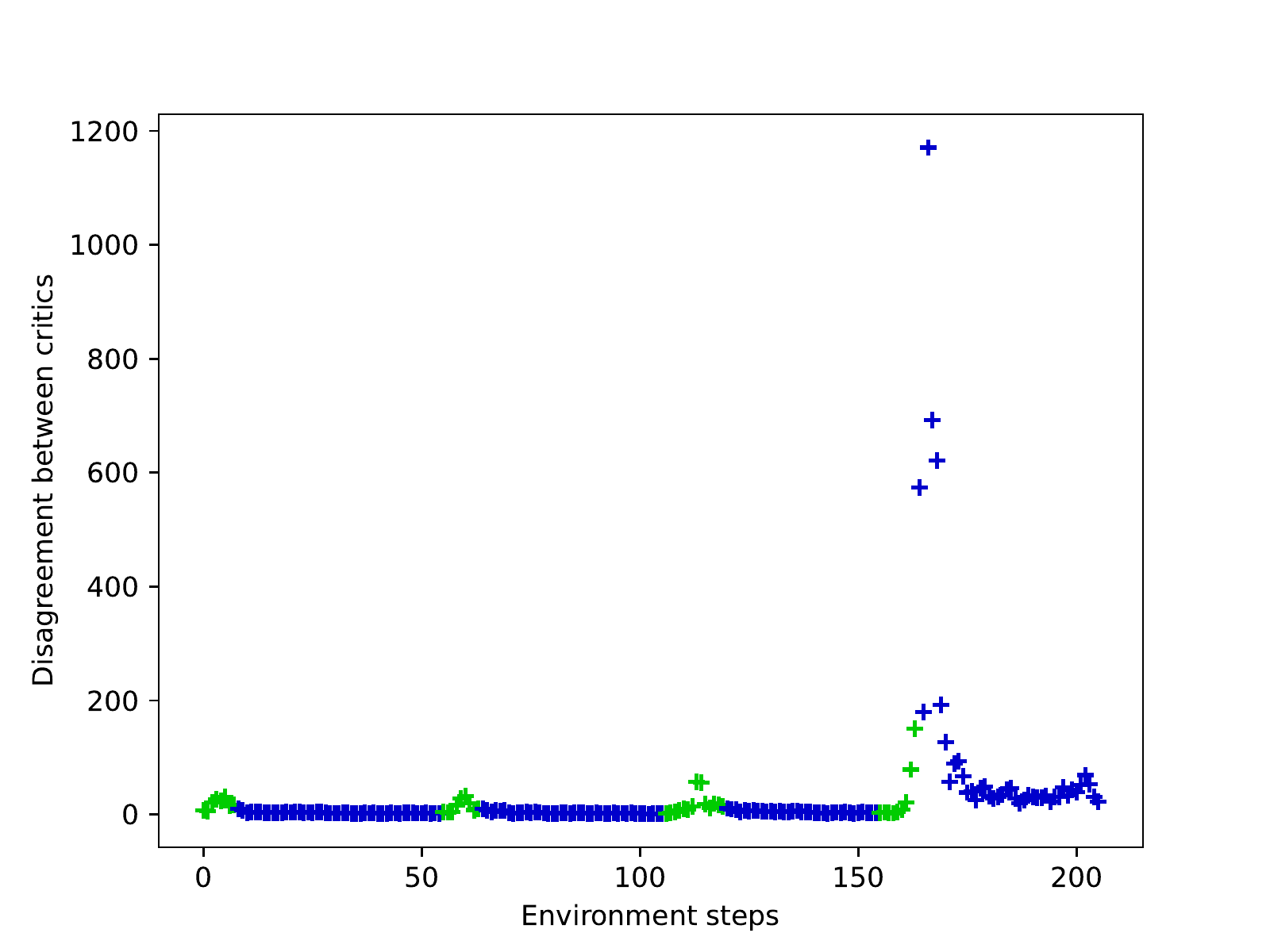}
    \caption{Critic disagreement in $L_1$ norm between the quantile critics of the 
    same ensemble $Z^{\piS}$ along a failed episode of the CartPoleGC.
    Dots are green when the GC policy is activated and blue when the safety policy is activated.}
    \label{fig:analyse_failures/921_scritic_critic_disagreement}
    \Description{Critic disagreement in $L_1$ norm between the quantile critics of the 
    same ensemble $Z^{\piS}$ along an episode of the CartPoleGC.}
\end{figure}

%\begin{wrapfigure}{r}{0.22\textwidth}
%\begin{center}
%    \includegraphics[width=0.21\textwidth]{images/cartpole_different_learn_seed_70_30/terminated_regret.pdf}
%\end{center}
%\caption{Occurrence of mistakes obtained during safe exploration for different pretraining seeds 
%with the CartPoleGC environment, $\thGCtoS = 70$ and $\thGCtoS = 30$}
%\Description{Occurrence of mistakes obtained during safe exploration for different pretraining seeds
%with the CartPoleGC environment, $\thGCtoS = 70$ and $\thGCtoS = 30$}
%\label{fig:pretrain_effect}
%\end{wrapfigure}
In this section, we propose to analyze the causes of the few mistakes we obtained with our
safe exploration method.
\refFig{fig:pretrain_effect} shows that the result strongly depends on pretraining. For one 
safety policy (policy 3), we obtain zero mistakes over the $3$ seeds of safe exploration training
while we obtain more than $25$ mistakes with the worst observed safety policy (policy 1). 
In order to avoid exhausting model selection engineering or cherry picking, we conducted a 
study to identify the failures' causes and obtain less variance with future algorithms.
Although one could decide to reduce thresholds, we do not want to penalize 
coverage because of simplistic choices.
Instead, we decided to analyse the trajectories stored in the episodic replay buffer $\buffer$ that ended
with a mistake to make a diagnosis. 
For CartpoleGC, with $(\thGCtoS, \thStoGC) = (70,30)$, 
all mistakes occur when the safety policy is active, meaning that the action selection mechanism
has seen the danger and switched too late from the GC to the safety policy. 
\refFig{fig:analyse_failures/921_scritic_critic_disagreement} shows a typical example of failure, 
characterized by a huge disagreement between critics before the mistake happens. This tends to
show that the safety policy has not been trained a lot on the corresponding states and actions. 
It also offers two straightforward complementary perspectives of research. The first is to take disagreement into 
account in the switching mechanism. The second is to finetune the safety policy during the safe exploration phase
to improve it, which is also confirmed by further analysis we left in the supplementary material.

% \begin{figure}[ht]
%     \includegraphics[width=0.25\textwidth]{images/cartpole_different_learn_seed_70_30/terminated_regret.pdf}
%     \caption{Mistakes occurrence obtained with the same safe exploration hyperparameters but with 
%     different pretrained policies, each associated to a seed.}
%     \label{fig:cartpole_different_learn_seed_70_30/terminated_regret}
%     \Description{Mistakes occurrence obtained with different seeds}
% \end{figure}

% \begin{figure}[ht]
%     \includegraphics[width=0.4\textwidth]{images/analyse_failures/921_safety_value_estimation.pdf}
%     \caption{}
%     \label{fig:analyse_failures/921_safety_value_estimation}
%     \Description{}
% \end{figure}

\section{Conclusion}

Our experiments demonstrate that the safe exploration framework we developed successfully trains a goal-conditioned policy while preventing mistakes during learning. We showed that incorporating a notion of distance to terminal states is crucial for safety and examined the impact of thresholds on both safety and goal-space coverage. As zero error over the course of safe exploration is our objective, we analyzed failure modes to identify ways to reduce mistakes. The primary cause of failures lies in insufficient pretraining of the safety policy in certain regions of the state space. Improvements could be achieved by considering disagreement between critics as an intrinsic motivation during pretraining or incorporating it into the action selection mechanism. 
Additionally, fine-tuning the safety policy during the safe exploration phase could expand the set of reachable states, further enhancing exploration. Also, we will focus on improving the evaluation procedure of the safety policy. Our experiments reveal that the key is not only to have a safety policy we trust but also to have a clear understanding of its limitations.

%Through our experiments, we have shown that the safe exploration framework we developed 
%is able to train a goal-conditioned policy and prevent mistakes while learning. 
%We have also shown that taking into account the estimated distance to terminal states is crucial to achieve 
%safety, and studied the effects of thresholds both on safety during exploration and goal-space coverage.
%As absolute zero error is virtually impossible, we conducted a study to explain failure modes
%and identify perspectives to reduce the number of mistakes. The main cause of failures appears to 
%be the pretraining, where the safety policy has not been trained enough on specific regions of the state 
%space. Taking into account disagreement between critics may lead to improvements, 
%as an intrinsic motivation during pretraining and/or as a new component of the action selection mechanism.
%Also one could finetune the safety policy during the pretraining phase. {\color{blue} AND WE WILL DO THAT IN FUTURE WORK}
%\textcolor{blue}{long term}

% \begin{acks}
% If you wish to include any acknowledgments in your paper (e.g., to 
% people or funding agencies), please do so using the `\texttt{acks}' 
% environment. Note that the text of your acknowledgments will be omitted
% if you compile your document with the `\texttt{anonymous}' option.
% \end{acks}

%%%%%%%%%%%%%%%%%%%%%%%%%%%%%%%%%%%%%%%%%%%%%%%%%%%%%%%%%%%%%%%%%%%%%%%%

%%% The next two lines define, first, the bibliography style to be 
%%% applied, and, second, the bibliography file to be used.

\bibliographystyle{ACM-Reference-Format} 
\bibliography{biblio}

%%%%%%%%%%%%%%%%%%%%%%%%%%%%%%%%%%%%%%%%%%%%%%%%%%%%%%%%%%%%%%%%%%%%%%%%

\newpage

\appendix

%\let\balance\relax
%\begin{multicols}{2}

\onecolumn

\section{Supplementary material}

\subsection{Videos}

\begin{itemize}
  \item \url{failure\_cartpole.mp4}: CartPoleGC failure mode described in the paper ($(\thGCtoS, \thStoGC) = (70, 30)$).
  \item \url{near\_bound\_cartpole.mp4}: CartPoleGC with a "difficult" goal near the environment bounds, thus near unsafe states ($(\thGCtoS, \thStoGC) = (70, 30)$).
  \item \url{simple\_cartpole.mp4}: CartPoleGC with a simple goal with $(\thGCtoS, \thStoGC) = (70, 30)$.
  \item \url{30\_30\_cartpole.mp4}: CartPoleGC with a simple goal with $(\thGCtoS, \thStoGC) = (30, 30)$ to show the impact of low thresholds on coverage.
  \item \url{skydiox2\_example.mp4}: Example of SkydioX2 reaching a goal ($(\thGCtoS, \thStoGC) = (10, 10)$)
\end{itemize}

Note that for CartPoleGC, the cart is blue when the safety policy is activated and green otherwise.
The goal is represented by a red box. 

\subsection{Further analysis of failure modes}

\begin{figure}[ht]
  \includegraphics[width=0.32\textwidth]{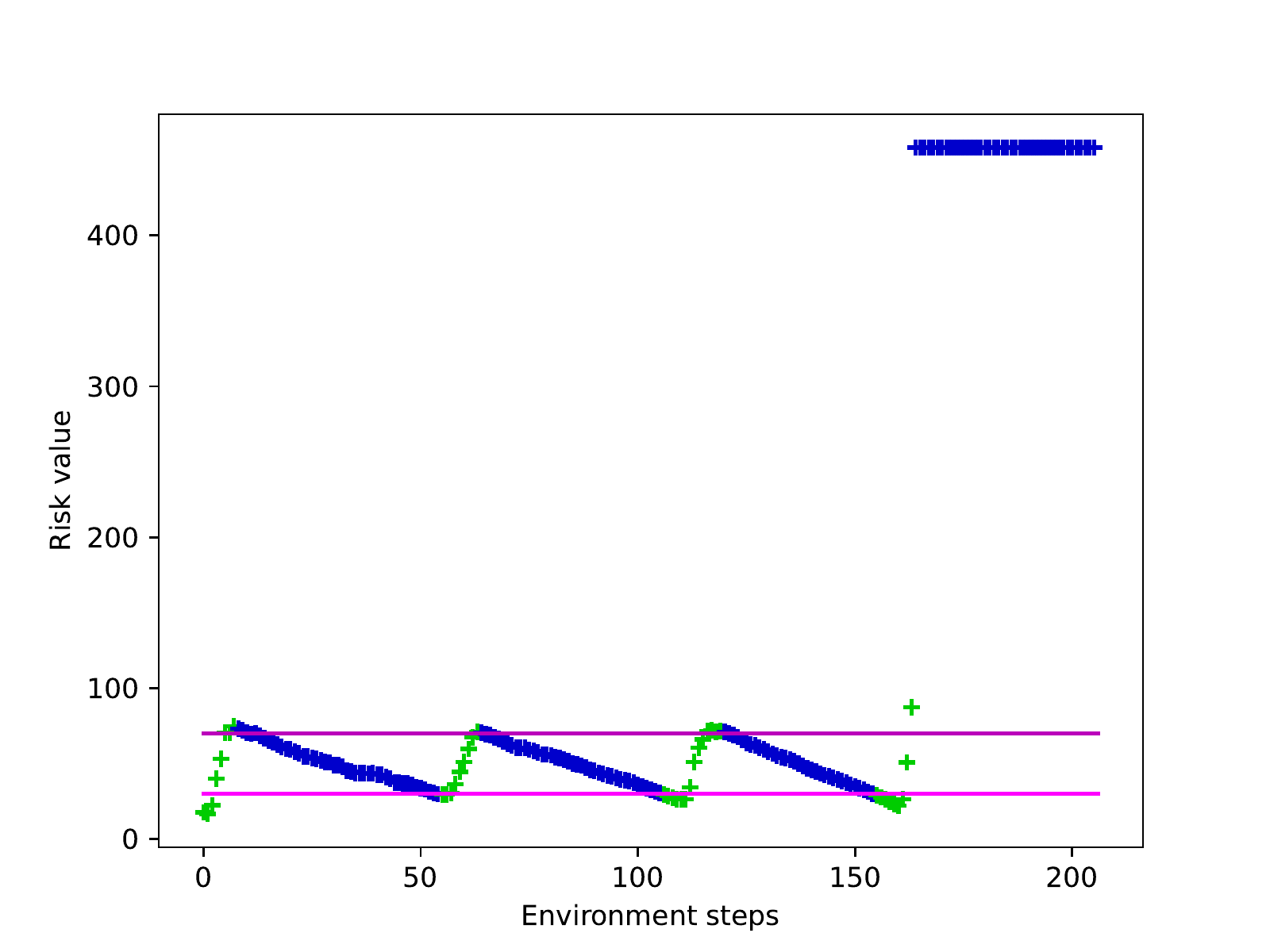}
  \includegraphics[width=0.32\textwidth]{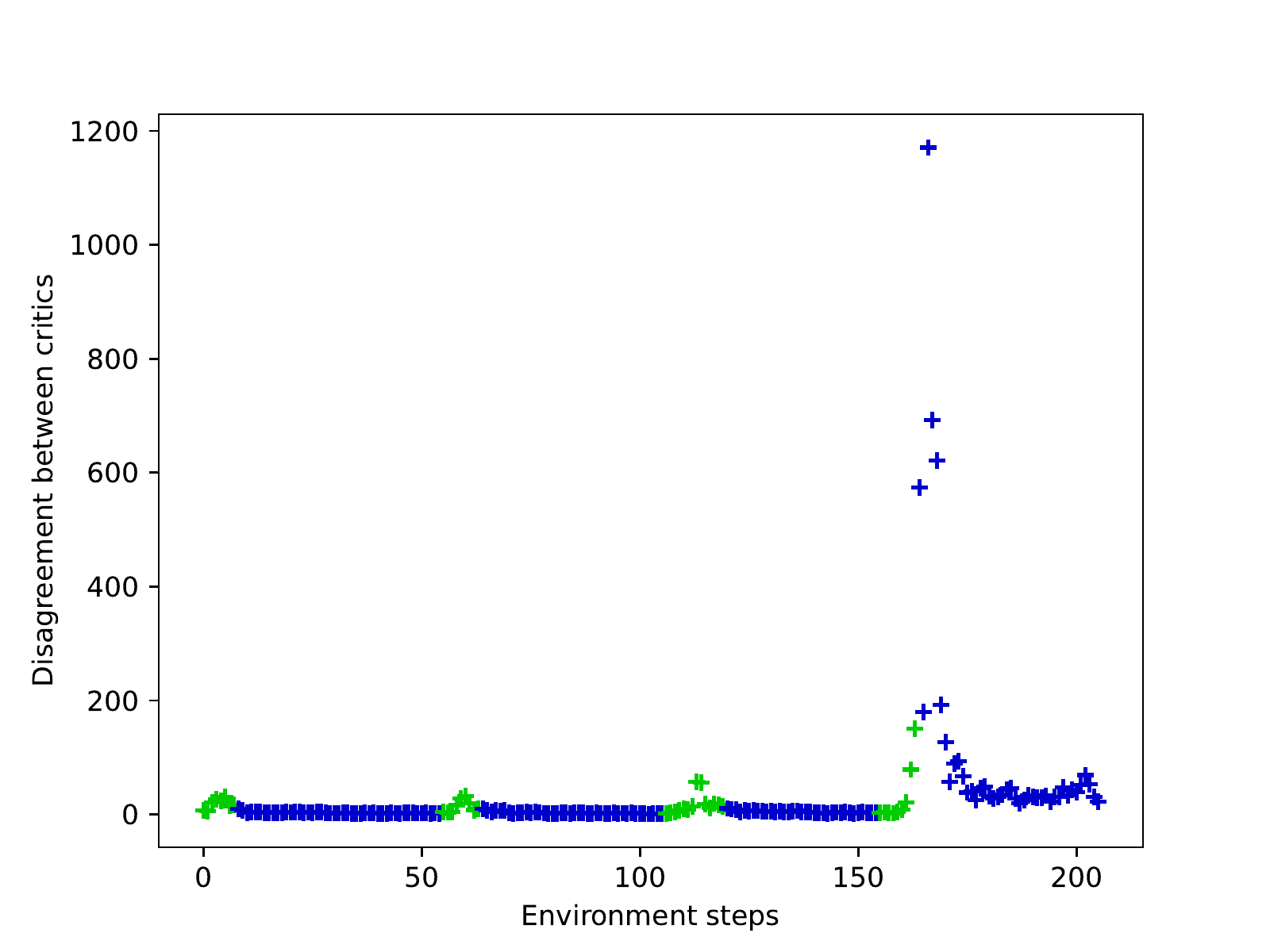}
  \includegraphics[width=0.32\textwidth]{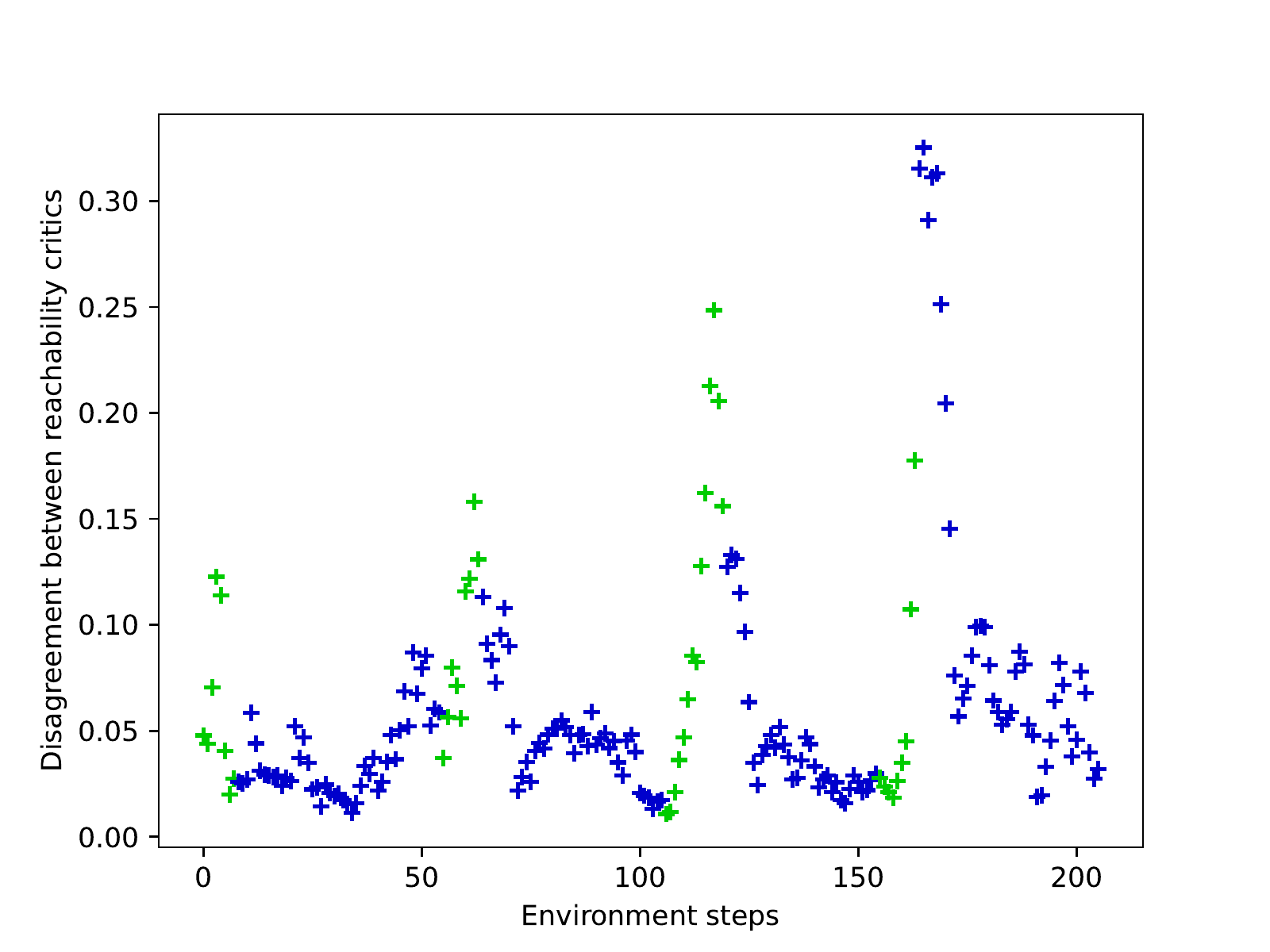}
  \caption{From left to right: Value of the risk function $\hat{\sigma}^{\piS}(s, a)$ along the failed episode
  shown as an example in Figure 10, where the thresholds are represented in magenta and purple ; Critic disagreement (Figure 10) ; Reachability critic disagreement.
  Blue dots correspond to the safety policy and green dots to the GC policy. One can see the hysteretic behavior on the left plot. A video of the failure (\url{failure\_cartpole.mp4}) is also attached.}
  \label{fig:failures}
  \Description{Further analysis of failure modes}
\end{figure}

We can see on the left plot (Figure \ref{fig:failures}) that the agent switches from the GC policy to the safety policy before making a 
mistake. 
We also observe that the oscillations of the disagreement, for both ensemble of critics,
are synchronized with the change of policies. Indeed, as the GC policy has a different objective
than the safety policy, it goes towards states that have been less visited by the safety policy during 
its pretraining. This phenomenon motivates safety finetuning for future works.

\newpage

\subsection{Goal-space coverage performance with safe exploration on CartPoleGC}

\begin{figure}[ht]
  \includegraphics[width=0.5\textwidth]{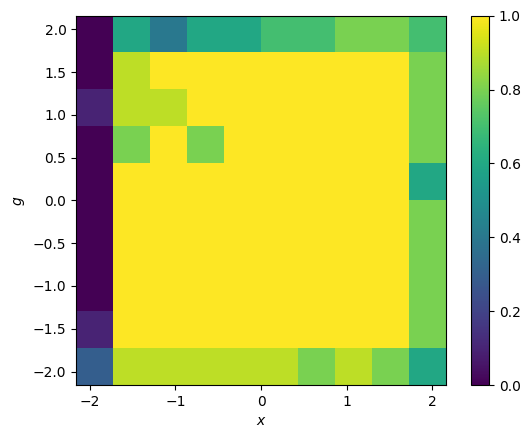}
  \caption{Coverage map obtained with $L\&S$ safe exploration variant and $(\thGCtoS, \thStoGC) = (70, 30)$
  on CartPoleGC. Only for this experience, the cartpole is reset on different $x$ positions. 
  Each cell corresponds to the combination of a starting position and a desired goal and we measure the 
  success rate. We can see that the success rate is lower for starting positions and goals near the 
  environment bounds, than for positions around the center. 
  The safety policy tends to prevent the agent from reaching goals near the bounds.
  In the same way, if the initial state is too close to the bound, the safety policy prevents the 
  GC policy from acting most of the time.}
  \label{fig:cartpole_cov}
  \Description{CartPoleGC coverage}
\end{figure}

% \newpage

\subsection{Constraint strategy}

\begin{figure}[ht]
\includegraphics[width=0.3\textwidth]{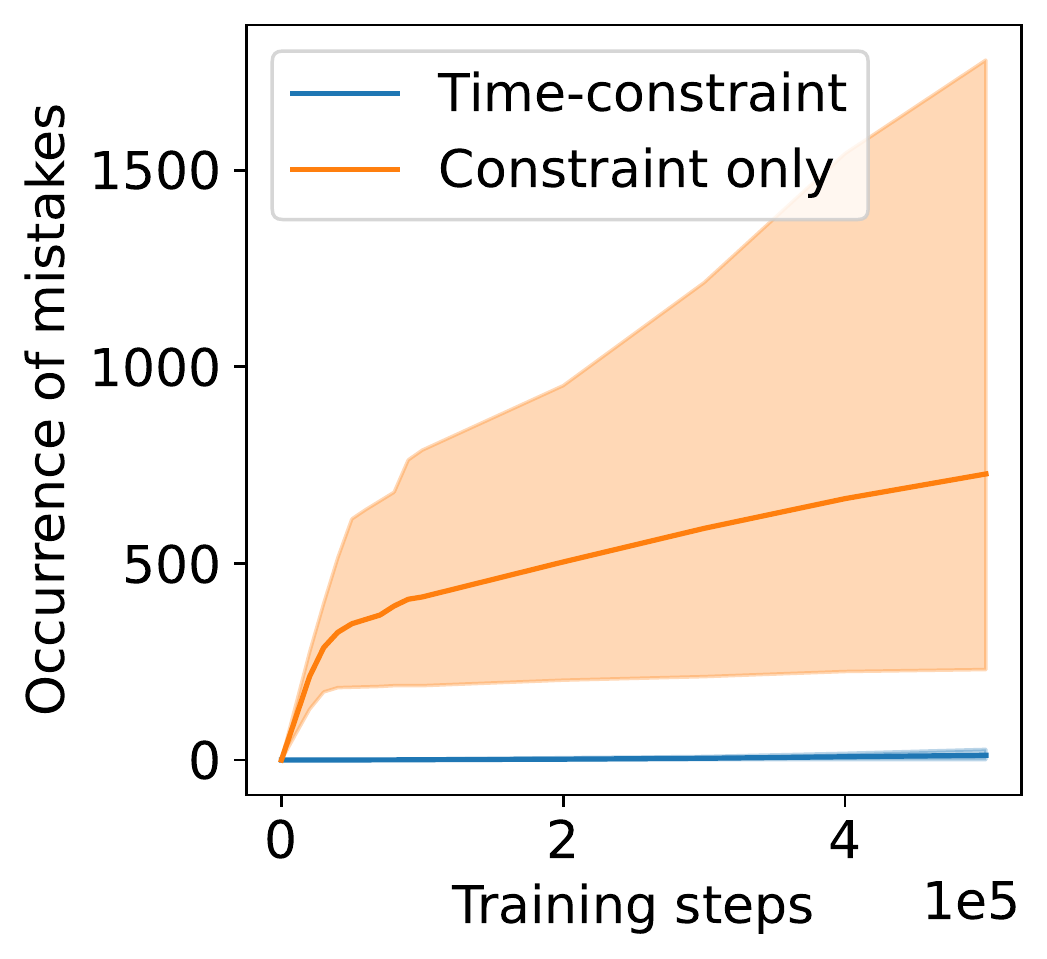}
\caption{Occurrence of mistakes obtained with the time-constraint strategy and the constraint (only) strategy 
on the CartPoleGC environment. Performance of constraint strategy in terms of safety is catastrophic.}
\label{fig:constraint_only}
\Description{Occurrence of mistakes obtained with the time-constraint strategy and the constraint (only) strategy 
on the CartPoleGC environment.}
\end{figure}

\subsection{Agent hyperparameters}

\begin{table}[th]
	\caption{Safety pretraining: TQC's hyperparameters}
	\label{tab:safety_pretraining_TQC}
	\begin{tabular}{rll}
    \toprule
		\textit{Name} & \textit{Value} \\ \midrule
		Actor learning rate & $3\times 10^{-4}$  \\
		Critic learning rate & $3\times 10^{-4}$ \\
		Temperature learning rate & $3\times 10^{-4}$ \\
    Initial temperature & $1.0$ \\
    $\tau$ & $5\times 10^{-3}$ \\
		No entropy backup & -  \\
		Discount factor & 0.99  \\ 
    Hidden layers & (256, 256) \\
    Number of critics & 5 \\
    Number of atoms per critic & 25 \\
    Number of quantiles to drop & 2 for CartPoleGC ; 0 for SkydioX2GC\\
    \bottomrule
	\end{tabular}
\end{table}

\begin{table}[th]
	\caption{Safety pretraining: Reachability critics' hyperparameters}
	\label{tab:safety_pretraining_RCRL}
	\begin{tabular}{rll}
    \toprule
		\textit{Name} & \textit{Value} \\ \midrule
		Critic learning rate & $3\times 10^{-4}$ \\
    $\tau$ & $5\times 10^{-3}$ \\
		Discount factor & 0.99  \\ 
    Hidden layers & (256, 256) \\
    Number of critics & 5 \\
    Number of atoms per critic & 25 \\
    \bottomrule
	\end{tabular}
\end{table}

\begin{table}[th]
	\caption{SAC and SAC-N hyperparameters for safe exploration}
	\label{tab:sac_sac_n}
	\begin{tabular}{rll}
    \toprule
		\textit{Name} & \textit{Value} \\ \midrule
		Actor learning rate & $3\times 10^{-4}$  \\
		Critic learning rate & $3\times 10^{-4}$ \\
		Temperature learning rate & $3\times 10^{-4}$ \\
    Initial temperature & $1.0$ \\
    $\tau$ & $5\times 10^{-3}$ \\
		No entropy backup & -  \\
		Discount factor & 0.99  \\ 
    Hidden layers & (256, 256) \\
    Number of critics (Specific to SAC-N) & 50 for CartPoleGC ; 10 for SkydioX2GC \\
    \bottomrule
	\end{tabular}
\end{table}

As for the buffer we choose to keep all transitions. There is no forgetting. 
Thus, the buffer size is always larger than the number of training steps.

%\end{multicols}

\end{document}